\documentclass[preprint,12pt]{elsarticle}

\let\today\relax
\makeatletter
\def\ps@pprintTitle{%
    \let\@oddhead\@empty
    \let\@evenhead\@empty
    \def\@oddfoot{\footnotesize\itshape
         {Accepted in Expert Systems with Applications} \hfill\today}%
    \let\@evenfoot\@oddfoot
    }
\makeatother

\usepackage{graphicx}
\usepackage{amssymb}

\usepackage{lineno}

\usepackage{todonotes}
\usepackage{amsmath}
\usepackage{subcaption}
\usepackage{url}
\usepackage[figuresright]{rotating}

\begin{document}

\begin{frontmatter}


\title{Sequence-based Dynamic Handwriting Analysis for Parkinson's Disease Detection with One-dimensional Convolutions and BiGRUs}



\tnotetext[t1]{This document is a collaborative effort.}

\author[a]{Moises~Diaz}
\ead{moises.diaz@atlanticomedio.es}
\author[b]{Momina~Moetesum} 
\ead{momina.buic@bahria.edu.pk}
\author[b]{Imran~Siddiqi} 
\ead{imran.siddiqi@bahria.edu.pk}
\author[c]{Gennaro~Vessio\corref{cor1}}
\ead{gennaro.vessio@uniba.it} 
   
\cortext[cor1]{Corresponding author}
\address[a]{Universidad del Atl\'antico Medio, Las Palmas de Gran Canaria, Spain}
\address[b]{Vision \& Learning Lab, Bahria University Islamabad, Pakistan}
\address[c]{Department of Computer Science, University of Bari ``Aldo Moro'', Italy}

\begin{abstract}
Parkinson's disease (PD) is commonly characterized by several motor symptoms, such as bradykinesia, akinesia, rigidity, and tremor. The analysis of patients' fine motor control, particularly handwriting, is a powerful tool to support PD assessment. Over the years, various dynamic attributes of handwriting, such as pen pressure, stroke speed, in-air time, etc., which can be captured with the help of online handwriting acquisition tools, have been evaluated for the identification of PD. Motion events, and their associated spatio-temporal properties captured in online handwriting, enable effective classification of PD patients through the identification of unique sequential patterns. This paper proposes a novel classification model based on one-dimensional convolutions and Bidirectional Gated Recurrent Units (BiGRUs) to assess the potential of sequential information of handwriting in identifying Parkinsonian symptoms. One-dimensional convolutions are applied to raw sequences as well as derived features; the resulting sequences are then fed to BiGRU layers to achieve the final classification. The proposed method outperformed state-of-the-art approaches on the PaHaW dataset and achieved competitive results on the NewHandPD dataset.
\end{abstract}


\end{frontmatter}


\section{Introduction}
Parkinson's disease (PD) is one of the most widespread and most disabling neurodegenerative disorders; it adversely affects the structure and functions of brain areas resulting in a gradual cognitive, behavioral, and functional decline~\citep{ascherio2016epidemiology}. At present, there is no cure, and the progressive deterioration of the patient can only be somehow managed during disease progression. Nevertheless, early diagnosis of PD could be crucial from the perspective of proper medical treatment to be administered as well as to evaluate the effectiveness of new drug treatments at prodromal stages~ \citep{bhat2018parkinson}. Moreover, the assessment of signs and manifestations of this specific disease is useful for its diagnostic differentiation from similar disorders, and for monitoring and tracking its progression as the disease advances. With this aim, over the years, a growing interest of the community has been observed in computer-aided diagnosis~\citep{parisi2018feature,ali2019early}. Such intelligent systems can effectively assist clinicians at the point of care, providing novel decision support tools, while reducing expenditure on public health.\\
 
Alterations in the brain caused by PD, such as neuronal loss, synaptic dysfunction, brain atrophy, etc., among others, can result in a malfunction of the motor system and its components. This is particularly manifested in performance impairment of previously learned motor skills. In this view, a unique role in the context of PD assessment can be safely assumed for handwriting. Handwriting is a complex activity involving perceptual-motor as well as cognitive components, the changes of which can be considered a promising \textit{biomarker} for disease assessment~\citep{de2019handwriting,vessio2019dynamic,2021_COGN_HandwTrends_Faundez}. Indeed, there is a growing body of knowledge which provides evidence that the automatic discrimination between unhealthy and healthy individuals can be accomplished through the use of simple and easy-to-perform handwriting tasks, e.g.~\citep{rosenblum2013handwriting,drotar2016evaluation,ammour2020new}. Developing a handwriting-based decision support tool is desirable, as it can provide a non-invasive, real-time, and low-cost solution to support the standard clinical evaluations carried out by human experts.\\
 
Within this research direction, on-line (\textit{dynamic}) systems based on the use of a digitizing tablet can be adopted. Such a device allows one to capture not only temporal and spatial variables of handwriting but also the pressure exerted by the pen over the writing surface, as well as measures of pen orientation and inclination. Moreover, this technology can acquire pen movement not only while the pen is in contact with the writing surface, but also when the pen is in the proximity of the surface, i.e.~``in-air''. Contrary to off-line (\textit{static}) features of handwriting, which can be analyzed after the writing process has already occurred, dynamic handwriting analysis deals with those features that can be acquired during the execution of the writing process. This can provide the system with rich dynamic information that can be exploited for disease diagnosis~\citep{drotar2014analysis}.\\

When designing such a system, a crucial step involves choosing the most appropriate features to describe handwriting. By directly feeding a classic statistical learning classifier with the time-series raw data, as acquired by the tablet, the model would suffer 
the burden of high dimensionality and thus overfitting. For this reason, several dynamic features have been derived from data in their raw form, ranging from traditional kinematic and spatio-temporal variables of handwriting to less common measures based, for example, on entropy and signal-to-noise ratio, e.g.~\citep{rosenblum2013handwriting,drotar2014decision,impedovo2019velocity}. It is worth noting that, to obtain complete statistical representations of the available features, mathematical functions of the feature vector (including mean, median, standard deviation, and so on) are generally computed. However, although this ``holistic'' approach can help the model find effective decision frontiers in the feature space, on the other hand, it may lose relevant information, as an arbitrarily long sequence is condensed into single-valued features.\\

Another approach to representing handwritten patterns is to use features automatically learned by deep learning models. Some recent works based on Convolutional Neural Networks (CNNs) address the automatic extraction of features from two-dimensional static images by exploiting dynamic information of the handwriting~\citep{pereira2018handwritten,diaz2019dynamically}. While this approach represents a robust alternative to manually engineered features, it also provides only a holistic view of the handwritten patterns under study; moreover, since it is black-box, this approach obfuscates the meaning of the features employed and their correlation with the concomitant disease.\\

An alternative way to process the time-series data without losing relevant information, which has not yet been explored to its full extent in this domain, is to apply the sequence-based neural learning paradigm using Recurrent Neural Network (RNN) models. The on-line recordings captured during writing can exhibit unique time-dependent patterns, which can be exploited to discriminate PD patients from healthy controls. Instead of compressing the original data into single-valued features, as done by many researchers, we want to exploit the sequential nature of the data to explicitly take time into account and gain new insights into the dynamic handwriting/drawing process. Although traditional methods have proved useful without explicitly modeling time, RNNs are powerful tools for modeling data with temporal or sequential structures of variable length~\citep{lipton2015critical}. Two commonly used recurrent units include the Long-Short Term Memory (LSTM)~\citep{hochreiter1997long} and Gated Recurrent Units (GRUs)~\citep{cho2014learning}. In recent years, systems based on these architectures have shown ground-breaking performance in traditionally challenging-to solve tasks, such as image captioning, language translation, and handwriting recognition, e.g.~\citep{you2016image,zhang2017drawing}.\\

The research presented in this paper represents a contribution to the state-of-the-art on sequence-based dynamic handwriting analysis for PD identification, extending our pilot study in this direction~\citep{Moetesum2020dynamic}. More specifically, we apply one-dimensional convolution to the raw sequences (as well as derived features), to take advantage of the abundant temporal information from the handwriting samples. This not only results in a robust feature representation, but also serves to sub-sample these sequences to mitigate overfitting while reducing training time. The resulting feature sequences are then fed to Bidirectional GRU (BiGRU) layers to achieve the final classification. This approach is best suited for capturing the temporal sequence of the handwritten patterns, in which muscle contractions and irregular movements due to Parkinsonism may be reflected. In fact, a significant improvement in the identification rate compared to the state-of-the-art is observed, in a fair experimental comparison carried out on the same dataset, namely PaHaW~\citep{drotar2016evaluation}, on a task-by-task basis. Moreover, an analysis of the significance of features is also carried out as a function of the classification rates reported. Finally, it is worth noting that PaHaW is based on samples acquired through a digitizing tablet. To further evaluate the robustness of our method, it was also tested on a dataset acquired via a smart pen, namely NewHandPD \citep{pereira2016deep}.\\

The rest of this paper is organized as follows. Section~\ref{sec:rw} reviews the notable works related to this problem. Section~\ref{sec:ma} and \ref{sec:me} respectively describe the materials and methods used in this research. Section~\ref{sec:r} reports and discusses the experimental results to highlight the effectiveness of the proposed method, while Section~\ref{sec:c} draws conclusions and presents the final remarks.

\section{Related Work}
\label{sec:rw}
In the context of Parkinson's disease assessment, dynamic handwriting analysis has been applied to investigate several issues and has attracted growing interest from diverse research areas (psychology, neuroscience, computer science, and so on). A large part of the literature on this topic investigated fine motor control impairments. The analysis of changes in handwriting facilitated the understanding of the brain-body functional relationships and led to some recognizable patterns of the sensorimotor dysfunction associated with PD, e.g.~\citep{teulings1991control,smits2014standardized,senatore2019paradigm}. Many other works, e.g.~\citep{eichhorn1996computational,randhawa2013repetitive,danna2019digitalized} have focused on studying the effects of medication on handwriting; analyzing the evolving patterns in handwriting can provide a useful tool for monitoring and tracking disease progression. More recently, significant research endeavors have been made towards the development of decision support tools to automatically discriminate between PD patients and healthy individuals~\citep{rosenblum2013handwriting,drotar2014analysis,zham2017distinguishing}. This research has been particularly stimulated by the recent advances in machine (deep) learning techniques. The ultimate goal is to provide clinicians with a complementary approach to their standard evaluation, which is fast, non-invasive, and low-cost. The present work is part of this research direction.\\

Among the notable contributions to PD identification from computerized handwriting analysis, the most significant series of works has been reported by Drot\'ar {\it et al.} All of their studies were carried out on the same dataset, i.e.~PaHaW~\citep{drotar2014analysis}, which was subsequently made available to the community. In~\citep{drotar2014analysis}, the authors investigated the extent to which classification performance can be improved, considering not only on-surface but also in-air movements as the two handwriting modalities appear to carry non-redundant information. In addition to computing conventional kinematic handwriting measures, such as velocity, acceleration, and jerk, \citet{drotar2014decision} also used relevant quantifiers based on entropy, signal energy, and empirical mode decomposition. These features provided novel insight and better understanding of the data. Subsequently, in~\citep{drotar2016evaluation}, the authors introduced additional fundamental features based on the pressure exerted over the writing surface. Specifically, they used the pressure values acquired by the tablet along with the rate at which the pressure signal changes over time.\\

The main factor contributing to the popularity of the PaHaW dataset in the research community is the collection of multiple handwriting tasks, ranging from the well-known Archimedes spiral drawing to word and sentence writing. Unfortunately, there are currently very few datasets freely available for research that provide multiple tasks, of varying degrees of complexity, performed by the same subjects. To better place our research in the literature panorama, we preferred to use this dataset. It is worth noting that, in all of the studies carried out by Drot\'ar {\it et al.}, the spiral task was undertaken without any significant impact on classification. This may have been due to the use of measures suitable only for handwriting; on the contrary, visual features, such as those extracted by Convolutional Neural Network models~\citep{diaz2019dynamically,moetesum2019assessing}, seem to overcome this issue.\\

\citet{impedovo2019velocity} improved the results obtained on the PaHaW dataset by combining classic features with new velocity-based features. The extended feature set includes parameters obtained from the Sigma-Lognormal model~\citep{ferrer2020idelog}, the Maxwell-Boltzmann distribution, and the Discrete Fourier Transform applied to the velocity profile of handwriting. \citet{rios2019analysis}, in addition to kinematic features, proposed to use geometrical and non-linear dynamic features. These features were proposed in the assumption that they are able to capture the irregularities of handwriting, which increase as the disease advances. In all the works discussed above, statistics computed on traditional hand-crafted dynamic features have been used to characterize PD.\\

Among other well-known contributions, \citet{pereira2016deep} introduced NewHandPD, a dataset of signals extracted from an electronic smart pen, which includes spiral and meander drawings. Each sensor of the pen outputs the overall signal acquired during the handwriting task, which can subsequently be represented as a time-series. The authors proposed to cast the problem of distinguishing PD from controls as an image recognition task through CNNs. Their strategy was to transform the signals provided by the smart pen into images. This research was one of the first applications of a deep learning-oriented approach to aid in the diagnosis of PD. The work was later extended in~\citep{pereira2018handwritten} and~\citep{afonso2019recurrence}. In~\citep{pereira2018handwritten}, CNNs were employed to learn texture-oriented features directly from the time-series-based images. The central hypothesis was that these features could encode hand tremors during handwriting. In~\citep{afonso2019recurrence}, on the other hand, the recurrence plot technique was used to map the pen signals into the image domain; these images were then fed into a CNN to learn effective features. A recurrence plot enables the visualization of repeated events of higher dimensions through projections onto low-dimensional representations and can be exploited to identify PD subjects.\\

Recently, in~\citep{diaz2019dynamically}, we proposed a ``dynamically enhanced'' representation of handwriting that consists of synthetically generated images obtained by jointly exploiting static and dynamic properties of handwriting. Specifically, we studied a static representation that embeds dynamic information based on drawing the points of the samples, instead of linking them, to preserve some velocity information, and adding pen-ups in the same way. The new handwriting representation, which was fed into CNNs to extract features automatically, was able to outperform the results obtained using static and dynamic handwriting separately on PaHaW. Unfortunately, although augmented with velocity and in-air information, the enhanced representation is still ``static'' and does not help the model reconstruct the temporal sequence of the handwriting movement.\\

More recently,~\citet{ribeiro2019bag} focused on the analysis of tremor, one of the most distinctive characteristics of PD. The authors proposed to learn temporal information from time-dependent signals by exploiting an RNN-based model along with an attention mechanism. The authors observed performance degradation due to long sequences, and the problem was addressed using a bag-of-sampling technique as a compact signal representation. Experimental results on the NewHandPD dataset compared favorably with the previous literature. The study advocated the potential of sequential data analysis for PD identification and motivated us to further explore this research direction.

\section{Materials}
\label{sec:ma}
Two datasets have been considered in this work. Both are publicly available for research and include time-based sequences of people with Parkinson's disease. Moreover, they both contain a similar number of specimens, which makes experimentation more balanced. First, the PaHaW dataset~\citep{drotar2014analysis} was used to adjust our system and find the best configuration. Second, the NewHandPD dataset~\citep{pereira2016deep} was used as an additional test bed for our method, as it contains handwriting samples acquired not through a tablet but via a smart pen. These data, which have not been seen by our system, would then lead to confirm the robustness of our system.

\subsection{PaHaW}
The ``Parkinson's disease handwriting database''  (PaHaW) collects handwriting data of 37 PD patients and 38 age and gender-matched healthy control (HC) subjects~\citep{drotar2014analysis}. Participants were enrolled at the First Department of Neurology, Masaryk University, and the St.~Anne's University Hospital, Brno, Czech Republic. All participants were right-handed, had completed at least ten years of education, and reported Czech as their native language. No significant between-group difference regarding age or gender was found. None of the subjects had a history or presence of any psychiatric symptom or disease affecting the central nervous system, with the exception of Parkinsonism in the PD group. Patients were only examined in their ON-state while taking dopaminergic medication, and, prior to acquisition, they were evaluated by a qualified neurologist. Additionally, the HC group underwent a thorough examination to ensure that no movement disorder or injury could have significantly affected handwriting.\\

All participants were asked to complete eight handwriting tasks following a pre-filled template:

\begin{enumerate}
    \item Drawing an Archimedes spiral;
    \item Writing in cursive the letter \textit{l};
    \item The bigram \textit{le};
    \item The trigram \textit{les};
    \item Writing in cursive the word \textit{lektorka} (``female teacher'' in Czech);
    \item \textit{porovnat} (``to compare'');
    \item \textit{nepopadnout} (``to not catch'');
    \item Writing in cursive the sentence \textit{Tramvaj dnes \v{u}z nepojede} (``The tram won't go today'').
\end{enumerate}

Since not all participants completed each task, we considered only those subjects who completed each of the eight tasks, i.e.~36 PD and 36 HC.\\

\begin{figure}[!ht]
\begin{subfigure}{.5\textwidth}
  \centering
  \includegraphics[width=.95\linewidth]{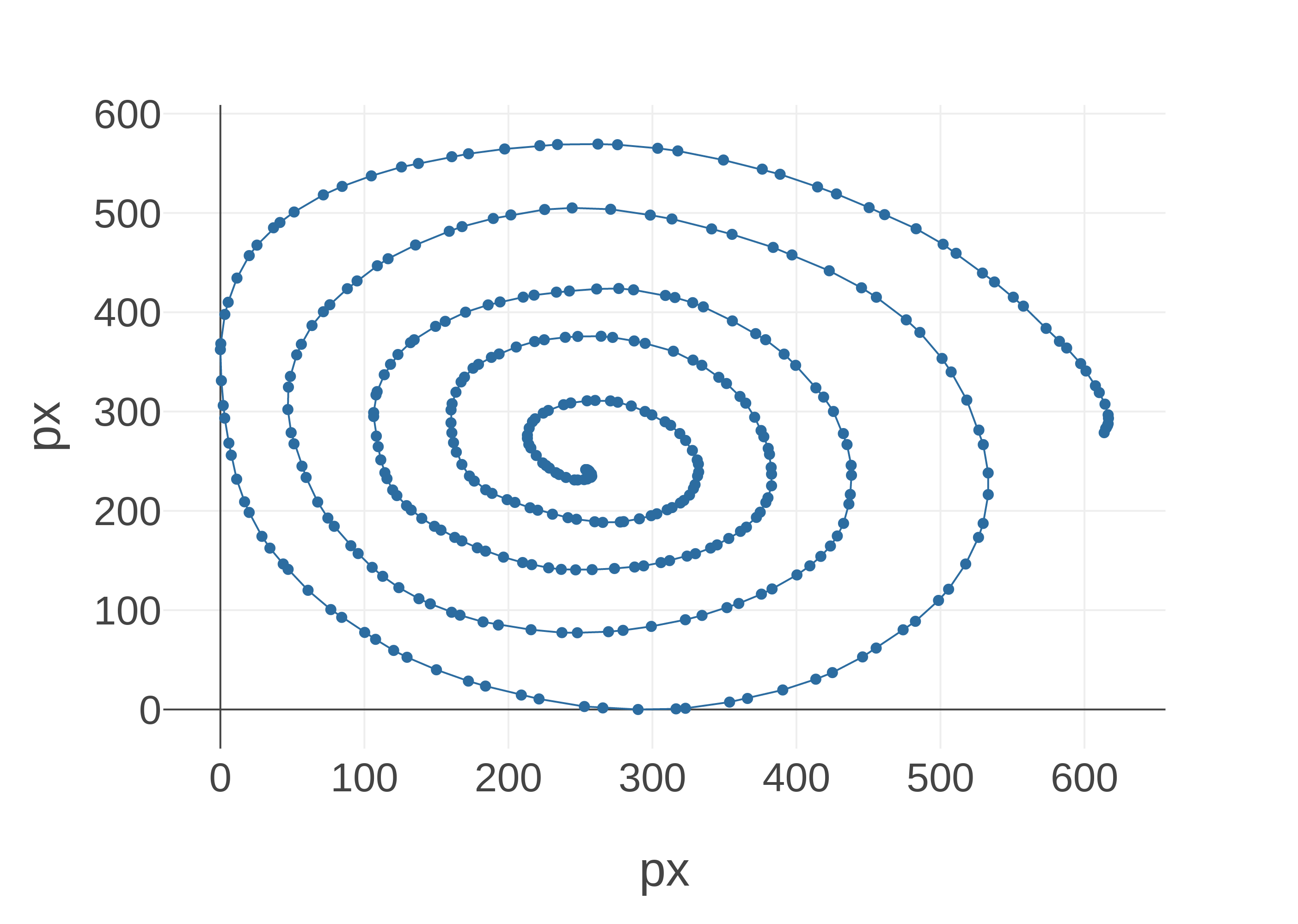}  
  \caption{Rendering for HC}
\end{subfigure}
~
\begin{subfigure}{.5\textwidth}
  \centering
  \includegraphics[width=.95\linewidth]{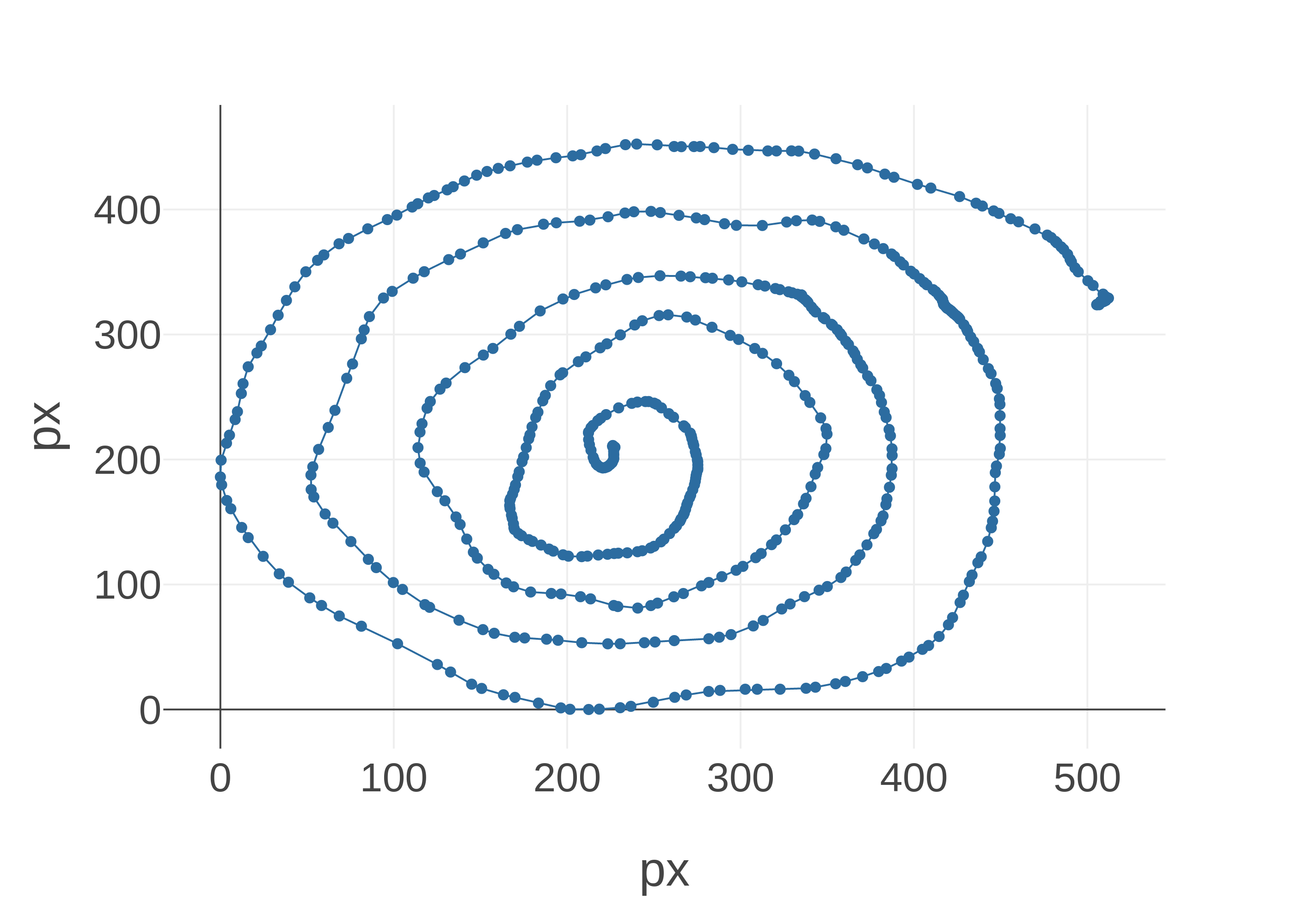}  
  \caption{Rendering for PD}
\end{subfigure}
\\
\begin{subfigure}{.5\textwidth}
  \centering
  \includegraphics[width=.95\linewidth]{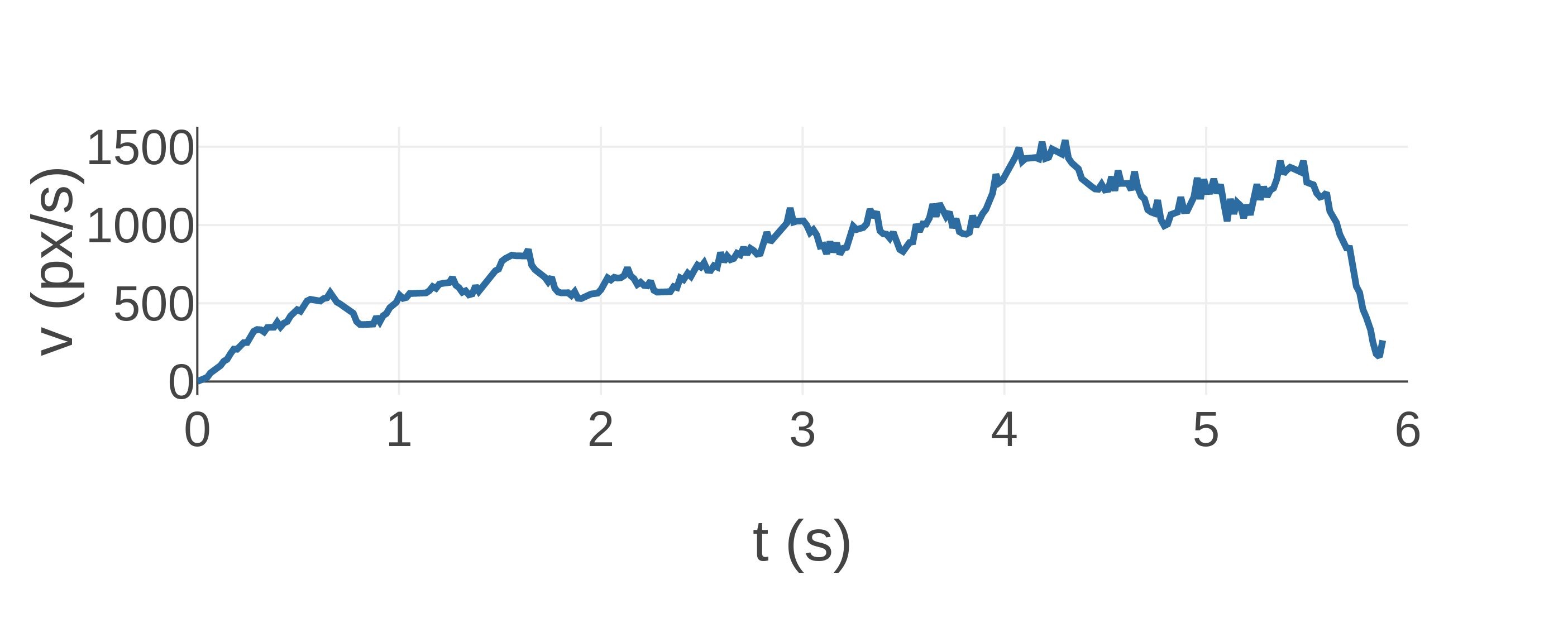}  
  \caption{Velocity for HC}
\end{subfigure}
~
\begin{subfigure}{.5\textwidth}
  \centering
  \includegraphics[width=.95\linewidth]{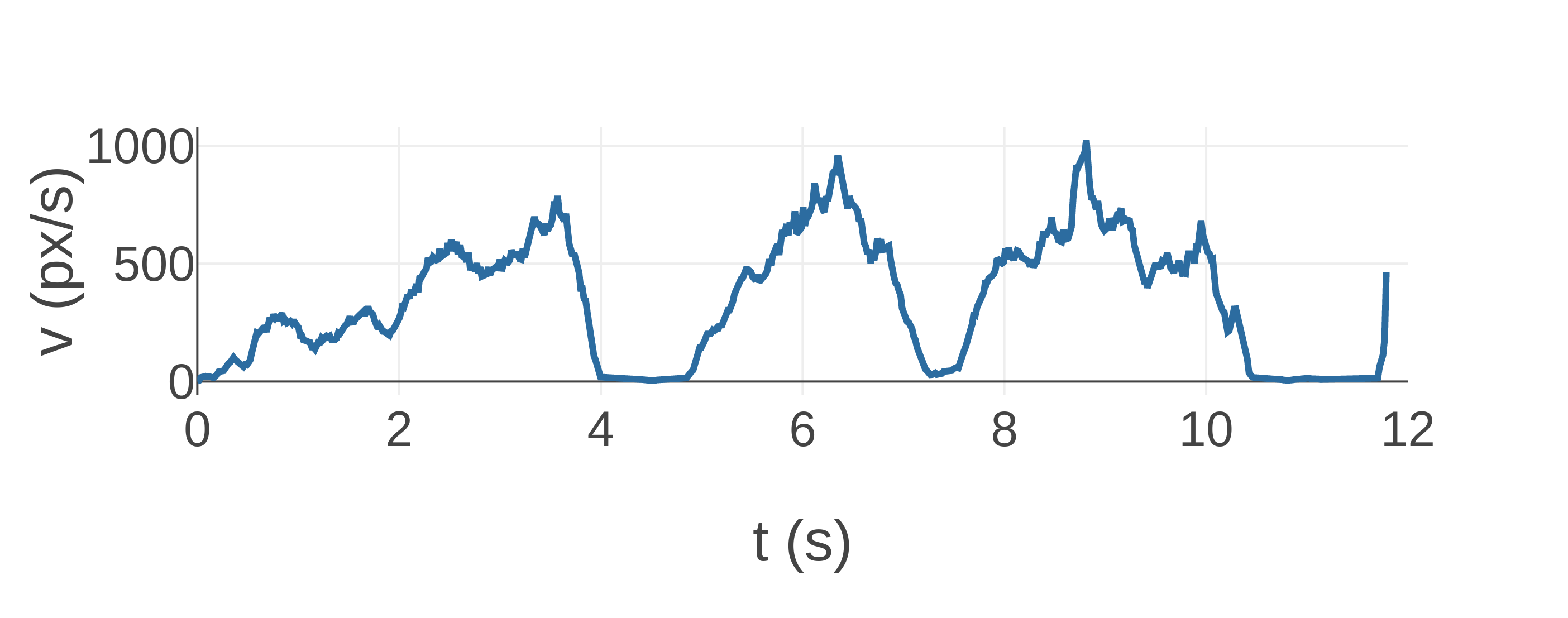}  
  \caption{Velocity for PD}
\end{subfigure}
\\
\begin{subfigure}{.5\textwidth}
  \centering
  \includegraphics[width=.95\linewidth]{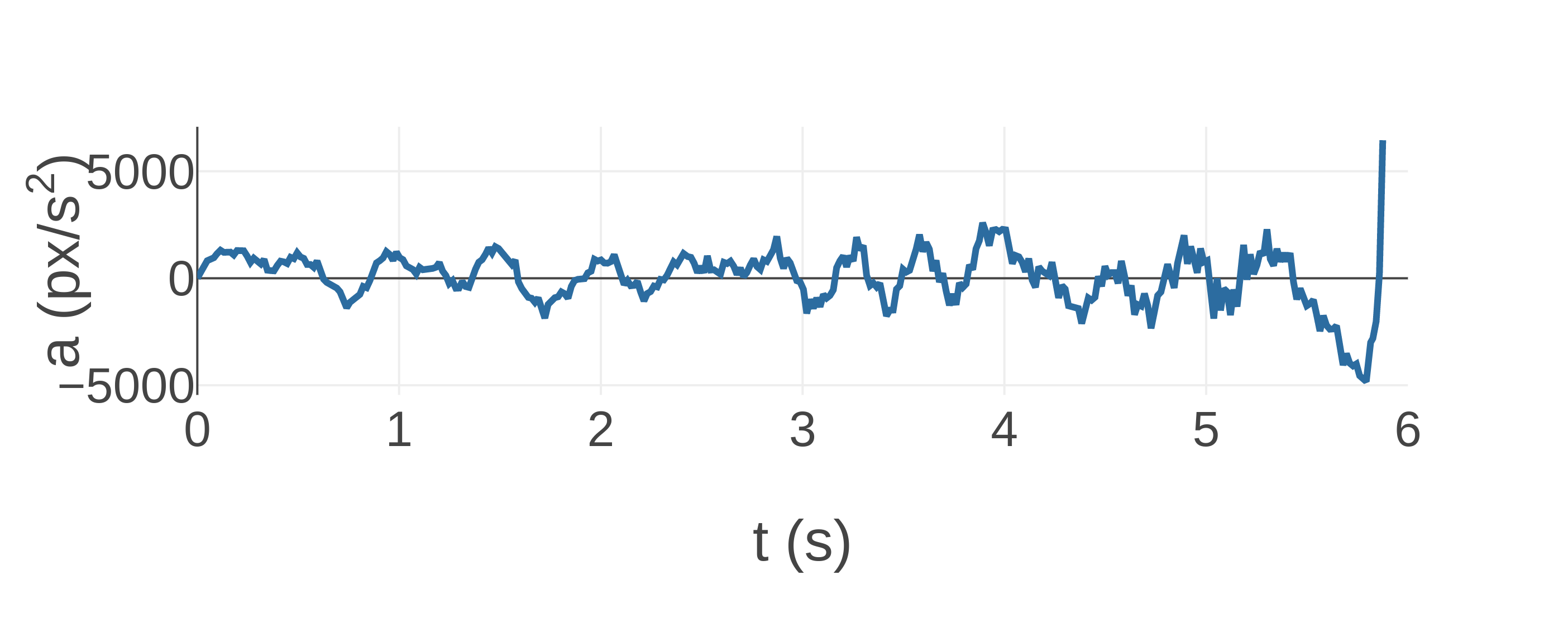}  
  \caption{Acceleration for HC}
\end{subfigure}
~
\begin{subfigure}{.5\textwidth}
  \centering
  \includegraphics[width=.95\linewidth]{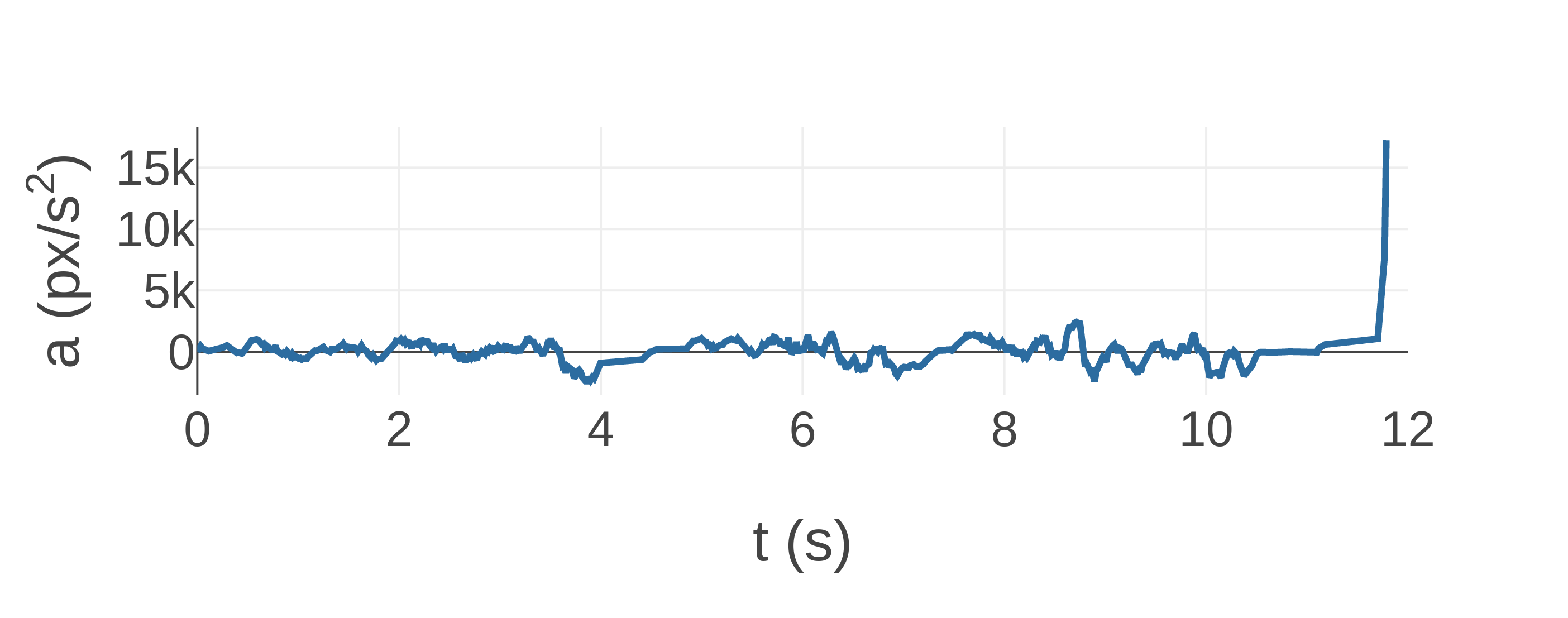}  
  \caption{Acceleration for PD}
\end{subfigure}
\\
\begin{subfigure}{.5\textwidth}
  \centering
  \includegraphics[width=.95\linewidth]{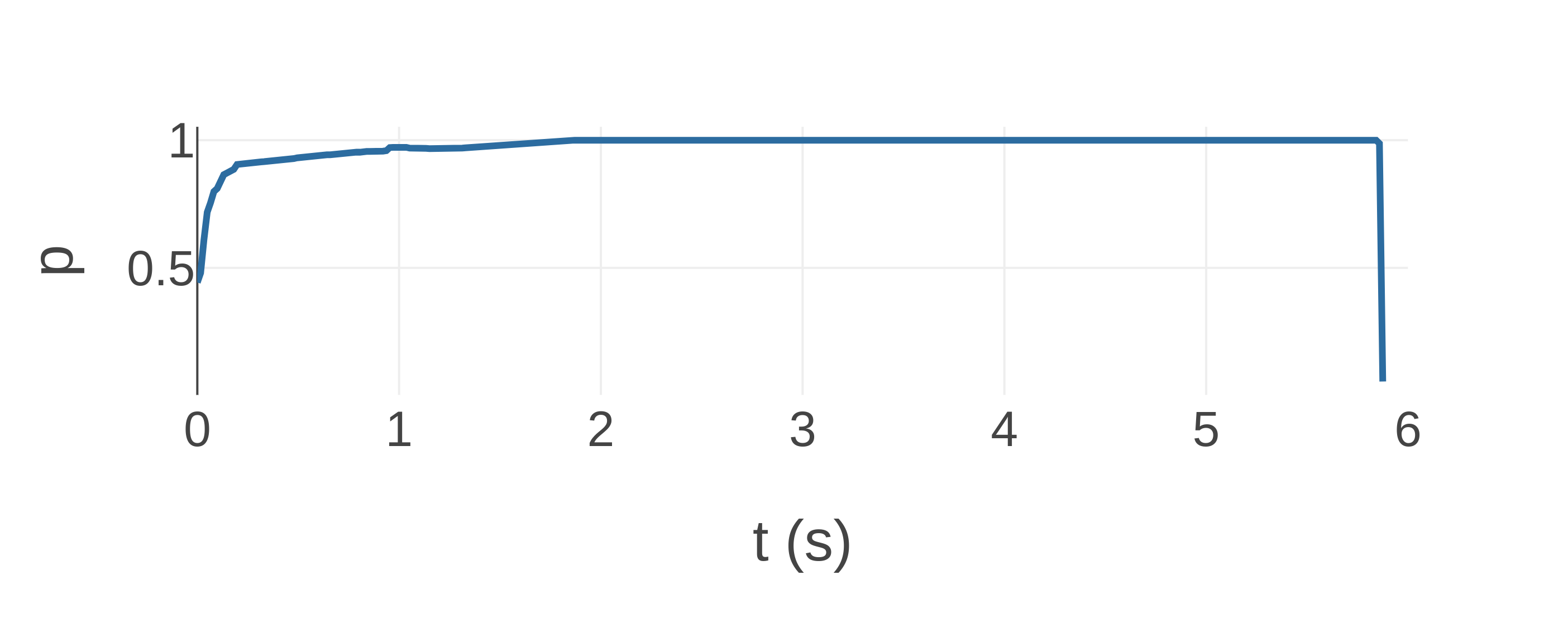}  
  \caption{Pressure for HC}
\end{subfigure}
~
\begin{subfigure}{.5\textwidth}
  \centering
  \includegraphics[width=.95\linewidth]{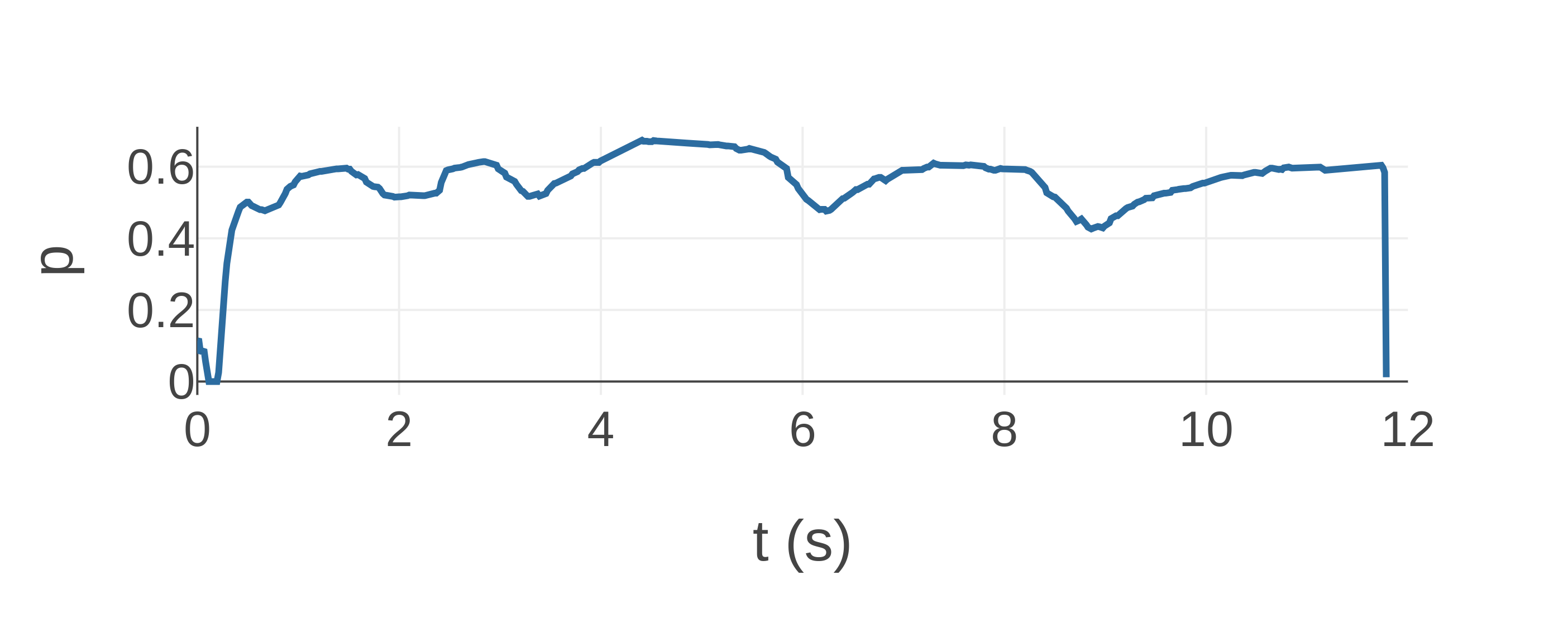}  
  \caption{Pressure for PD}
\end{subfigure}
\caption{Archimedes spiral task performed by a healthy adult (on the left) and a Parkinsonian patient (on the right). From top to bottom are the rendered task, and the velocity, acceleration and pressure profile (the curves are plotted as a function of advancement during writing). A doctor can typically only see the handwritten pattern left on the paper. Dynamic handwriting features, on the other hand, reveal a lot more information. Parkinsonian handwriting, in fact, shows irregularities and fluctuations that are not shown by the healthy handwriting. (Note that these data do not belong to PaHaW or NewHandPD, but were acquired during the HAND project~\citep{angelillo2019performance}).}
\label{fig:samples}
\end{figure}

The handwriting signals were recorded using a Wacom Intuos digitizing tablet, overlaid with a blank sheet of paper. Like many other professional tablets, the raw data acquired are the $x$- and $y$-coordinates of the pen tip, the corresponding time stamps, measures of pen inclination, i.e.~tilt-$x$ and tilt-$y$, and pen pressure. The button status is also available, which is a binary variable with value 0 for pen-ups (``in-air movement'') and 1 for pen-downs (``on-surface movement''). The sampling rate was 200 samples per second. Few sample images of healthy and Parkinsonian writing are depicted in Fig.~\ref{fig:samples}.

\subsection{NewHandPD}


The NewHandPD database~\citep{pereira2016deep} is an extension of the previous HandPD corpus~\citep{pereira2016new}. The first database consisted of images from two drawing tasks, i.e.~the typical spiral cognitive test and a modified spiral (``meander'') test performed by 
healthy individuals and people with Parkinson's disease. However, the new corpus, NewHandPD, contains both offline images and online signals (time-based sequences) of the two groups. The handwriting signals were acquired through a technology other than a tablet, i.e.~an electronic smart pen (BiSP).\\

Specifically, NewHandPD contains images and dynamic data from 31 patients and 35 healthy people. The gender of the participants was fairly balanced (39 males and 29 female), while most of them were right-handed writers (59 of 66 participants). They were asked to complete a handwriting-based test consisting of the following 12 exams:
\begin{enumerate}
    \item Four tasks related to spirals;
    \item Four tasks related to meanders;
    \item Two circled movements (one in-air and another on-surface);
    \item Two diadochokinesis tests (one with the left hand and the other with the right).
\end{enumerate}
    
The electronic smart pen recorded the following temporal data in its six channels for each exam:
\begin{enumerate}
    \item Microphone;
    \item Finger grip;
    \item Axial pressure of ink refill;
    \item Tilt and acceleration in $x$ direction;
    \item Tilt and acceleration in $y$ direction;
    \item Tilt and acceleration in $z$ direction.
\end{enumerate}



\section{Methods}
\label{sec:me}
\begin{figure}[!ht]
    \centering
    \includegraphics[width=0.95\textwidth]{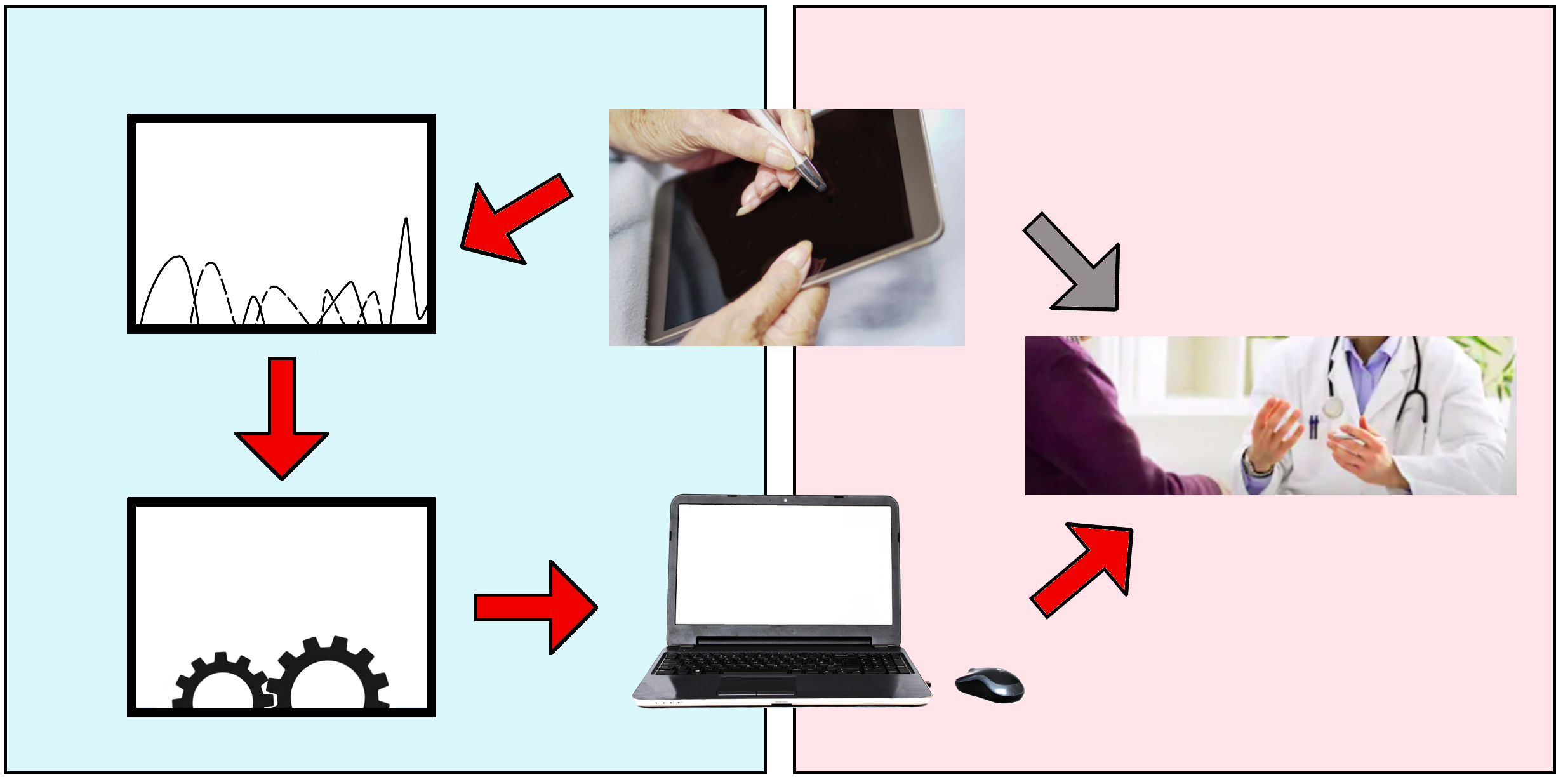}
    {\footnotesize \put(-110,30){Objective input for}
    \put(-110,17){medical diagnosis}
    \put(-110,140){Traditional subjective}
    \put(-89,127){input for medical}
    \put(-55,114){diagnosis}
    \put(-135,168){\underline{\small \bf  APPLICATION}}
    \put(-307,168){\underline{\small \bf SYSTEM}}
    \put(-205,54){P(PD) = }
    \put(-185,44){\bf 0.34}
    \put(-335,54){Deep learning}
    \put(-329,44){architecture}
    \put(-325,148){Sequences}
    \put(-335,138){from dynamic}
    \put(-330,128){handwriting}
    }
    \caption{Schematic overview of the proposed system. The data of different handwriting tasks performed by the same subject are acquired through a digitizing tablet. The handwriting specimen of a single task is represented as a multi-dimensional vector of several dynamic features to be provided to a deep learning machine. The system output estimates whether or not the sample belongs to the Parkinsonian class.}
    \label{fig:workflow}
\end{figure}
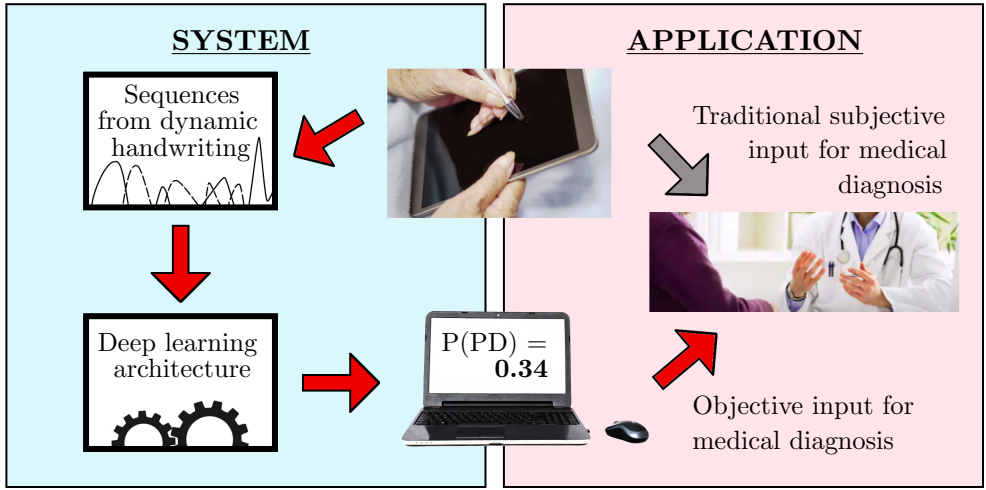

This section introduces our proposed methodology to exploit the potential of sequential information hidden in the time-series handwriting signals for automatic PD identification. Traditionally, medical diagnosis is based on subjective observations from different series of clinical tests. In our study, a computer-aided procedure is proposed to exploit non-visual information for such tests. As an additional element to the medical diagnosis, an objective result is provided, which outperforms the state-of-the-art.  Figure~\ref{fig:workflow} depicts a schematic workflow of the proposed method, while details are provided in the following.

\subsection{Input Features}
It is assumed that raw time-series data sampled from a conventional digitizing tablet are available: pen position, time stamp, pen pressure, pen inclination, and button status. Kinematic and pressure features can be derived from these raw measures. Kinematic features include the tangential, horizontal and vertical displacement, velocity, acceleration, and jerk. Displacement is the straight-line distance between two consecutive sampled points:

\begin{equation*}
d_i = \sqrt{(x_i-x_{i-1})^2+(y_i-y_{i-1})^2},
\end{equation*}

\noindent where $i=2, \ldots, Z$ (where $Z$ is the number of sampled points), and $d_1=0$. Given the typically high sampling rate of the acquisition device, it generally provides a good approximation of the actual pen trajectory. From this measure, velocity, acceleration, and jerk can be calculated straightforwardly as the first, second, and third derivative of displacement, respectively. This feature set can be enriched by (separately) considering displacement, velocity, acceleration, and jerk along the horizontal and vertical directions. Additionally, to use the pressure data, together with the raw value, we also calculated the first derivative of pressure, which represents the rate of change of pressure over time. 
An overview of the input features we have considered is provided in Table~\ref{tab:features}.\\

These features are suitable for our classification problem, as several studies, e.g.~\citep{broderick2009hypometria,smits2014standardized}, reported alterations of Parkinsonian handwriting in terms of writing time, writing size, applied pressure, and velocity fluctuations. Note that we do not consider other commonly used spatio-temporal variables, such as stroke size and duration, overall time, etc., as they are expressed as a single-valued feature rather than a time-dependent vector feature~\citep{drotar2014analysis}.\\

\begin{table}[!t]
\scriptsize
\centering
\begin{tabular}{l c l}
\hline
\textbf{Feature} & \textbf{\textit{r/d}} & \textbf{Description}\\
\hline
$x$ & \textit{r} & $x$-coordinate of the pen position during handwriting \\
$y$ & \textit{r} & $y$-coordinate of the pen position during handwriting \\
Pressure & \textit{r} & Pressure exerted over the writing surface \\
Tilt-$x$ & \textit{r} & Angle between the pen and the surface plane \\
Tilt-$y$ & \textit{r} & Angle between the pen and the plane vertical to the surface \\
Button status & \textit{r} & Boolean variable indicating whether the pen is on-surface or in-air \\
Displacement & \textit{d} & Pen trajectory during handwriting \\
Velocity & \textit{d} & Rate of change of displacement with respect to time \\
Acceleration & \textit{d} & Rate of change of velocity with respect to time \\
Jerk & \textit{d} & Rate of change of acceleration with respect to time \\
Horizontal/vertical displacement & \textit{d} & Displacement in the horizontal/vertical direction \\
Horizontal/vertical velocity & \textit{d} & Velocity in the horizontal/vertical direction \\
Horizontal/vertical acceleration & \textit{d} & Acceleration in the horizontal/vertical direction \\
Horizontal/vertical jerk & \textit{d} & Jerk in the horizontal/vertical direction \\
First derivative of pressure & \textit{d} & Rate of change of pressure with respect to time\\
\hline
\end{tabular}
\caption{Dynamic handwriting features. Abbreviations: \textit{r} = raw feature; \textit{d} = derived feature.}
\label{tab:features}
\end{table}

Each handwriting sample $S_n$ can therefore be represented as a multidimensional vector of $m$ dynamic features, where each feature $X_i$ consists of a sequence of $T$ time-steps:

 \begin{equation*}
 \begin{split}
     S_{n=1}^N &= \{ X_1^n, X_2^n, \ldots X_m^n \} \\
X_{i=1}^m &= \{ x_i^{t_1}, x_i^{t_2}, \ldots x_i^{T} \},
\end{split}
 \end{equation*}

\noindent where $N$ is the size of the dataset. The length of the sequential data recorded by the tablet can be arbitrarily long and depends on the time taken by the subject, as well as on the task performed. Since the input sequences can be of varying length for each sample, we fix the time-step length in the pre-processing step. Relatively long time sequences can negatively affect training time, while concise features can lead to underfitting. On the basis of the available data, we first propose to compute the average length of the overall sequences and then to use this length as a cut-off. When the sequences are shorter than the cut-off length, zero-padding is added.

\subsection{One-dimensional Convolutions}
The time-dependent sequences are fed into one-dimensional (1D) convolutional layers with stride greater than 1. The advantage of employing 1D convolution is two-fold. First, these layers sub-sample the input sequences, thereby reducing the overall training cost of the RNN model. Second, 1D convolutions can extract local temporal information from the input sequences, thus performing a pre-training step towards learning meaningful temporal dependencies. Combining 1D convolutions and RNNs is beneficial, especially when dealing with very long sequences that would hardly be processed with an RNN. In our case, we can have a few thousands time-steps for a sequence. The effect of the convolutional layers is to turn the long input sequence into much shorter (down-sampled) pieces of higher-level, locally invariant features. The sequence of extracted features then represents the input for the RNN component of the network.\\

Several filters of varying sizes are used in each layer to extract information across multiple time-scales. In particular, we employed two convolutional layers in cascade. The first applies $8$ filters, with a kernel of size $5$ and stride $5$. The following layer involves a higher number of $16$ filters with a reduced kernel of size $3$ and stride $3$. A commonly used ReLU nonlinearity follows both layers. It is common practice to augment the number of filters in the following layers as the low-level features of the previous one can be combined in several ways to obtain higher-level representations.

\subsection{Bidirectional Gated Recurrent Units}
Recently, deep learning models such as recurrent neural networks have gained popularity in sequential data analysis~\citep{yu2019review}. Unlike a conventional feed-forward neural network, an RNN has a recurrent hidden state $h_i^t$, whose activation at a given time $t$ depends on the previous state at time $t-1$. This is shown in the following equation:


\begin{equation*}
    h_i^t = g \left( W . x_i^t + U. h_i^{t-1} + b \right),
\end{equation*}

\noindent where $W$ and $U$ are weight matrices, $b$ is the bias term, $x_i$ is the input vector and $g$ the activation function. Despite their effectiveness in modeling sequential data, RNNs are known to suffer from the vanishing gradient problem due to which they may fail to capture long-term dependencies. To address this issue, element-wise non-linearities are typically adopted, which employ two types of recurrent units: the Long-Short Term Memory (LSTM) and the Gated Recurrent Unit (GRU). Although both variants can improve performance, we have chosen to use a GRU-based model as GRUs are less computationally expensive than LSTMs due to the lower number of gates and therefore fewer parameters to learn. \\


To further enhance learning, we propose to use Bidirectional GRU (BiGRU) layers. In a BiGRU, two independent GRUs are combined in a bidirectional fashion, with one reading the input sequence in the forward direction. Conversely, the other reads the same sequence in the backward direction. The hidden states from each GRU are then concatenated, as shown in the following:

\begin{equation*}
\begin{split}
\left( h_i^t \right )_f &= GRU_f \left( x_i^t,h_i^{t-1} \right ), \: \: \forall t \in \left[ 1, T_i \right ] \\
\left( h_i^t \right )_b &= GRU_b \left( x_i^t,h_i^{t-1} \right ), \: \: \forall t \in \left[ T_i, 1 \right ] \\ 
h_i^t &= \left[ \left( h_i^t \right )_f; \left( h_i^t \right )_b \right ], 
\end{split}
\end{equation*}

\noindent where $f$ and $b$ stand for \textit{forward} and \textit{backward}, respectively. We then process a sequence in both directions to capture patterns that a unidirectional model might overlook.\\

\begin{sidewaysfigure}
    \includegraphics[width=1.0\columnwidth]{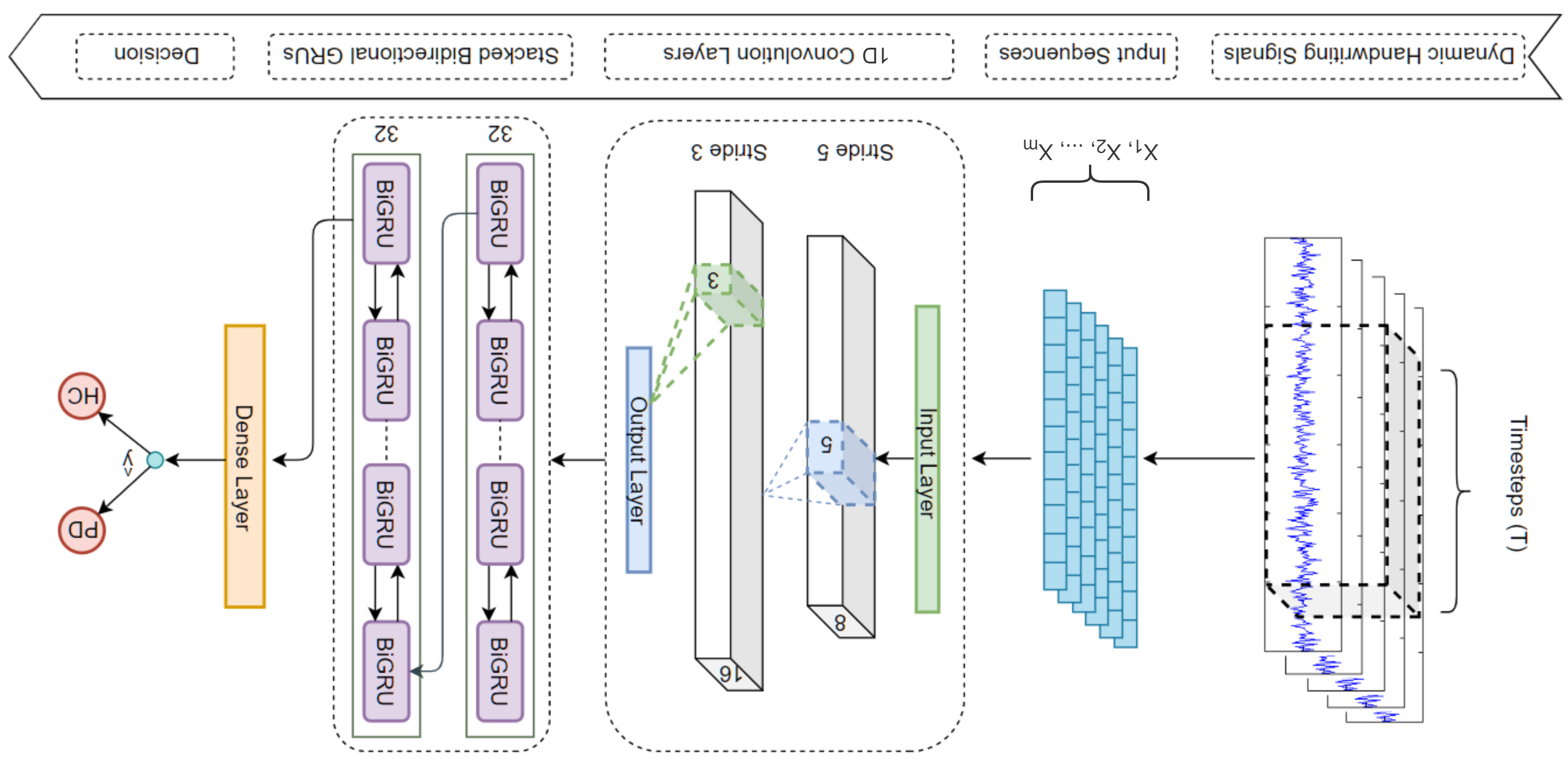}
    \caption{Architectural details of the convolutional-BiGRU model for PD/HC prediction. Raw data from handwriting signals are converted into feature sequences. The input to the model is a sequence of length $m$ and each time-step is a vector of features listed in Table~\ref{tab:features}. There are two 1D convolutional layers with $8$ and $16$ filters and strides of $5$ and $3$, respectively. Two Bi-directional GRU layers, each with 32 units, follow. The output layer has a single neuron with sigmoid activation to predict one of the two classes (PD or HC).}
    \label{fig:model}
\end{sidewaysfigure}

Two BiGRU layers (32 hidden units each) are stacked on top of the previously mentioned convolutional layers. The two BiGRU layers are interleaved by a conventional dropout and a recurrent dropout, both with a dropout rate of $0.1$, to further mitigate overfitting. The output of the last BiGRU is finally sent to an output neuron, with a sigmoid activation attached. Since the problem to be solved can be modeled as a binary classification task (PD/HC), the overall network is required to minimize a classical binary cross-entropy loss function. Training was done using back-propagation with the Adam optimizer and a learning rate of 0.001 on randomly sampled mini-batches of size $16$. The overall combined Convolutional-BiGRU model is illustrated in Figure~\ref{fig:model}.

\section{Experiments}
\label{sec:r}
In this section, we report the results of a series of experiments aimed at assessing the effectiveness of the proposed method:

\begin{itemize}
    \item The first experiment evaluated the predictive potential of the proposed system on the PaHaW dataset and ascertained the contribution of the individual subsets of features to the overall classification accuracy;
    
    \item The second experiment fairly compared, on the same dataset and with the same validation scheme, the proposed method with state-of-the-art approaches to PD detection through dynamic handwriting analysis;
    
    \item The third experiment consisted of ablation studies aimed at justifying some architectural choices we made for the construction of the model; 
    
    \item Finally, the fourth experiment evaluated the model with the best configuration on the NewHandPD dataset to further validate its robustness on data acquired through a slightly different technology.
\end{itemize}

In the following, the mean accuracy values are reported, averaged over all the iterations of a 10-fold cross-validation scheme. This validation strategy is usually preferred when the size of the data is small. 
Moreover, for the best model, we also report classification performance in terms of area under the ROC curve (AUC), sensitivity, and specificity, which are commonly used in diagnostic settings.

\subsection{Classification Results on PaHaW}
Table~\ref{tab:classification} summarizes the mean accuracy values reported by the proposed model by varying the feature set given as input. Excellent performance is observed in the overall \textit{derived} feature set, including kinematic as well as pressure features, calculated from the raw input acquired by the tablet. The highest predictive potential achieved a mean accuracy of over 90\% in almost all cases. In contrast, the overall \textit{raw} feature set reported the lowest classification rates. Not surprisingly, the kinematic features, which contribute most to the aforementioned derived feature set, exhibit the top second accuracy for all tasks among the individual feature groups. These results highlight the effectiveness of these features in capturing the impairments Parkinsonian patients have as they typically do not write with the same constancy as healthy subjects, showing a lower writing speed, with continuous acceleration peaks, e.g.~\citep{kotsavasiloglou2017machine,jerkovic2019analysis}. Significantly lower results, on the other hand, are obtained with pressure features, if considered alone. Patients generally apply less pressure on the writing surface; moreover, the pressure signal assumes erratic values due to muscular difficulties~\citep{rosenblum2013handwriting}. However, pressure is generally considered controversial in the literature, especially from the perspective of signature verification~\citep{linden2018dynamic, diaz2019perspective}, as results differ among studies. Another interesting observation is that pen inclination resulted in relatively better performance, with mean accuracy of more than 80\% in two cases. The pen angle information is typically discarded in most related studies. The present results indicate that pen inclination can also be exploited in addition to kinematic and pressure information to further enrich feature representation. It is important to recall that all of these features are first fed to a series of convolutional layers, so the final set of sequences that is provided to the recurrent layers is expected to be a rich and robust representation of the discriminating attributes between PD subjects and healthy controls. These findings also corroborate the hypothesis that sequence learning may be preferred to holistic approaches for PD detection through the dynamics of handwriting.\\

\begin{table}[!t]
\scriptsize
\centering
\begin{tabular}{l c c c c c}
\hline
\textbf{Task} & \textbf{Raw} & \textbf{Inclination} & \textbf{Pressure} & \textbf{Kinematic} & \textbf{Derived}\\
\hline
Spiral & 70.36\% & 63.39\% & 76.25\% & 85.00\% & 93.75\% \\
\textit{lll} & 67.50\% & 87.68\% & 74.46\% & 93.75\% & 96.25\% \\
\textit{le le le} & 71.25\% & 78.39\% & 72.68\% & 92.50\% & 88.75\% \\
\textit{les les les} & 69.11\% & 79.11\% & 65.54\% & 88.75\% & 90.00\% \\
\textit{lektorka} & 63.93\% & 65.54\% & 61.07\% & 90.00\% & 93.75\% \\
\textit{porovnat} & 61.96\% & 73.21\% & 68.57\% & 91.07\% & 91.25\% \\
\textit{nepopadnout} & 69.11\% & 78.75\% & 67.68\% & 88.57\% & 92.50\% \\
Sentence & 65.89\% & 80.71\% & 60.89\% & 95.00\% & 92.50\% \\
\hline
\end{tabular}
\caption{Classification performance of the proposed method. The contribution of individual subsets of features is shown for each task.}
\label{tab:classification}
\end{table}

Further considerations from Table~\ref{tab:classification} can be drawn by looking at the results obtained task by task. In general, the different feature sets agree that \textit{lll}, \textit{les les les} and the sentence are among the most discriminating tasks. No-sense words composed of one or more character repetitions have been used frequently for PD assessment, e.g.~\citep{bidet2011handwriting,smits2014standardized}, showing the impairment of Parkinsonian patients in fine motor control during loop-like movements. Indeed, PD patients may produce slower and more irregular movements; moreover, they may write letters in a more segmented fashion, showing micrographia over time when writing. Recently, \citet{senatore2019paradigm} 
found that Parkinsonian writing during a familiar \textit{l}-shape movement is characterized by a lack of fluency, slowness, and abrupt changes of direction. These difficulties support the hypothesis that the fine-tuning of the motor plan involved is deteriorated due to PD while executing a writing task.\\

The importance of the sentence task, already observed in~\citep{drotar2014decision}, is also confirmed in our experiments. In fact, writing a long sentence can require a greater cognitive load, particularly a high degree of simultaneous processing. Therefore, it can increase the effects of the disease on handwriting. The high degree of simultaneous processing is due to several reasons, including the involvement of linguistic skills, attention, and memory. Producing loop-like movements and writing a sentence offers the opportunity to better evaluate the motor plan between one character or word and the next. In fact, a hesitation or pause between two characters or words can highlight the need to re-plan the writing activity. Conversely, fluid writing reveals the presence of early motor planning~\citep{bidet2011handwriting, 2020_LognormalityChapter_Diaz}. In particular, a sentence allows one to capture a large number of in-air movements between components; conversely, a single word could be written without leaving the pen from the writing surface~\citep{drotar2016evaluation}. \\

Another observation concerns the spiral task. As mentioned above, the task was undertaken without any significant impact on classification in previous studies, e.g.~\citep{drotar2014decision}. Instead, similar to what we previously observed~\citep{diaz2019dynamically,moetesum2019assessing}, the Archimedes spiral task has achieved high classification accuracy here for almost all feature sets. This reinforces the clinical validity of the task, as clinical experts commonly use it for screening for early signs of PD. One reason may be due to the time it takes to complete the task, as spiral drawing requires continuous on-surface strokes in all directions and, therefore, can better capture changes in the dynamics of handwriting in all directions.\\

\begin{figure}[!ht]
\begin{subfigure}{.5\textwidth}
  \centering
  \includegraphics[width=.85\linewidth]{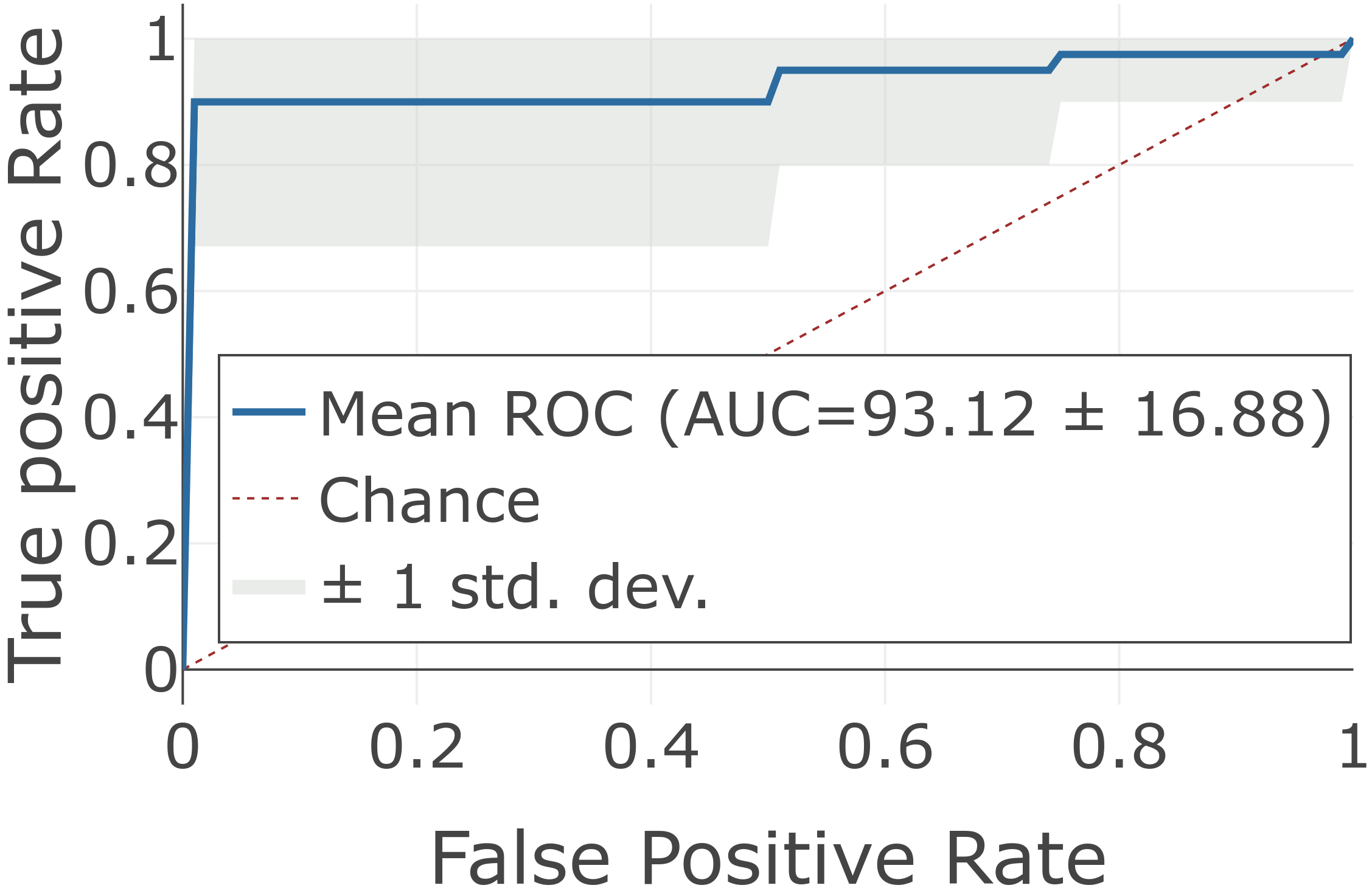}  
  \caption{ROC plot for Task 1}
\end{subfigure}
~
\begin{subfigure}{.5\textwidth}
  \centering
  \includegraphics[width=.85\linewidth]{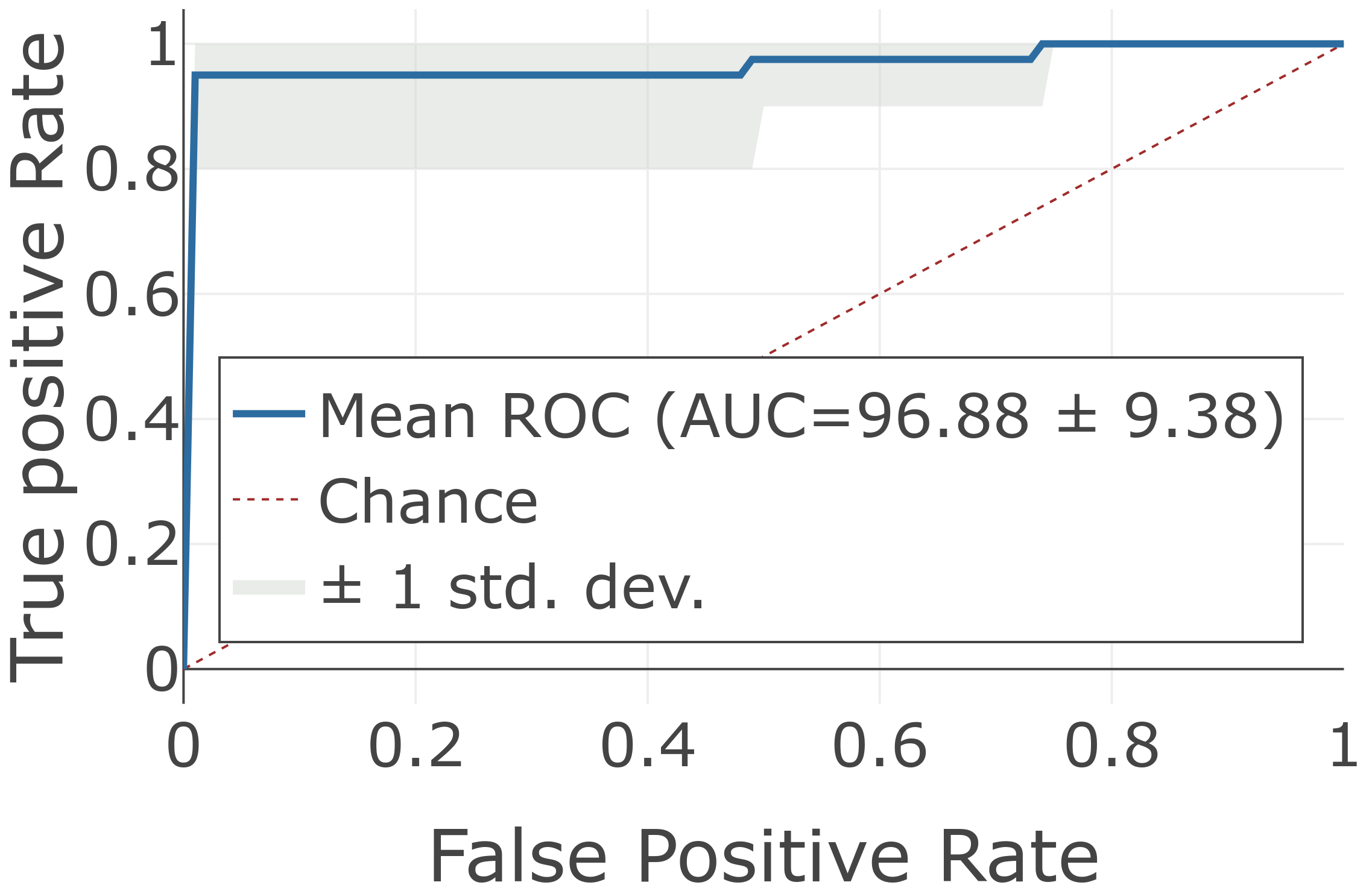}  
  \caption{ROC plot for Task 2}
\end{subfigure}
\\
\begin{subfigure}{.5\textwidth}
  \centering
  \includegraphics[width=.85\linewidth]{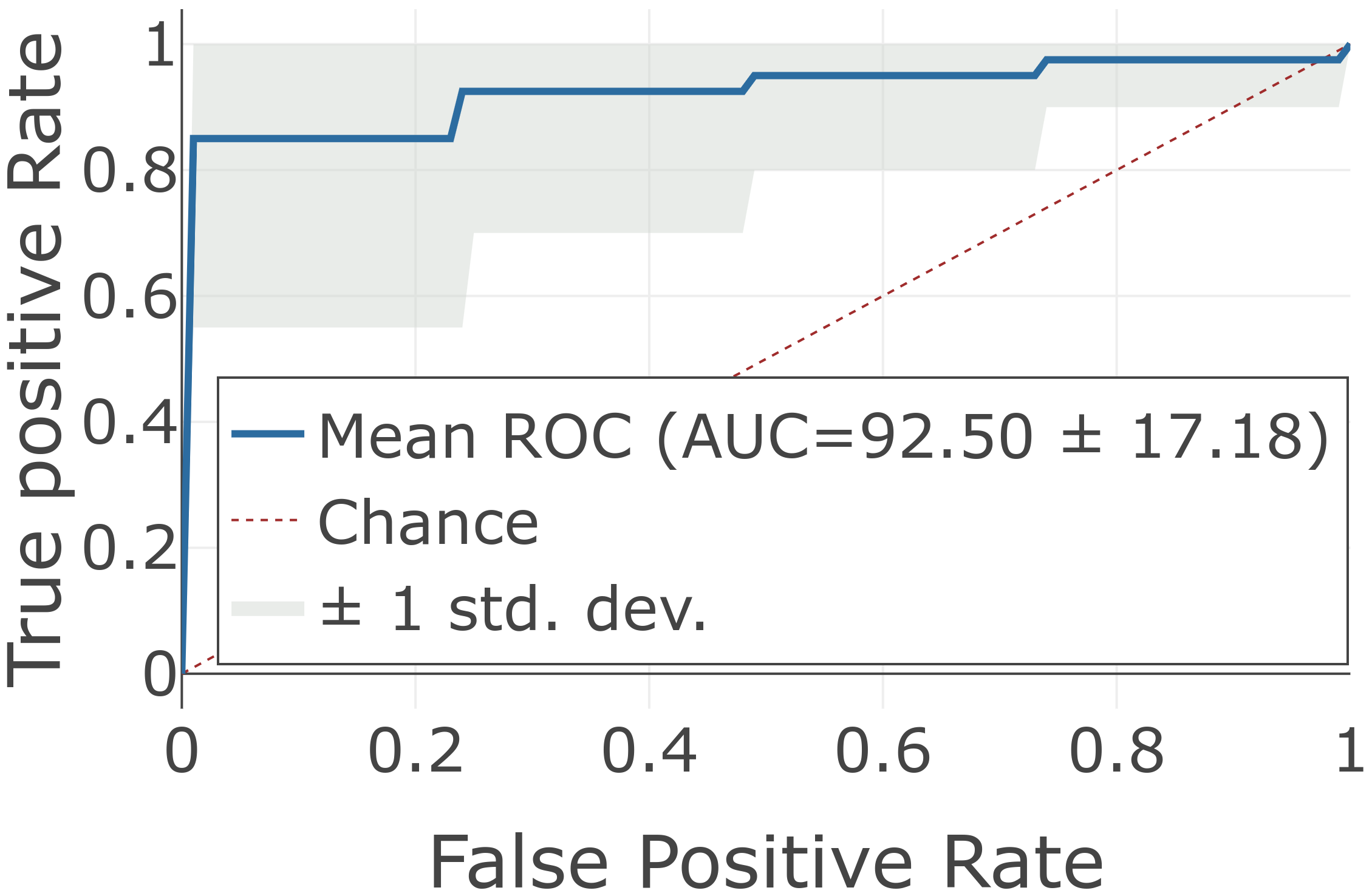}  
  \caption{ROC plot for Task 3}
\end{subfigure}
~
\begin{subfigure}{.5\textwidth}
  \centering
  \includegraphics[width=.85\linewidth]{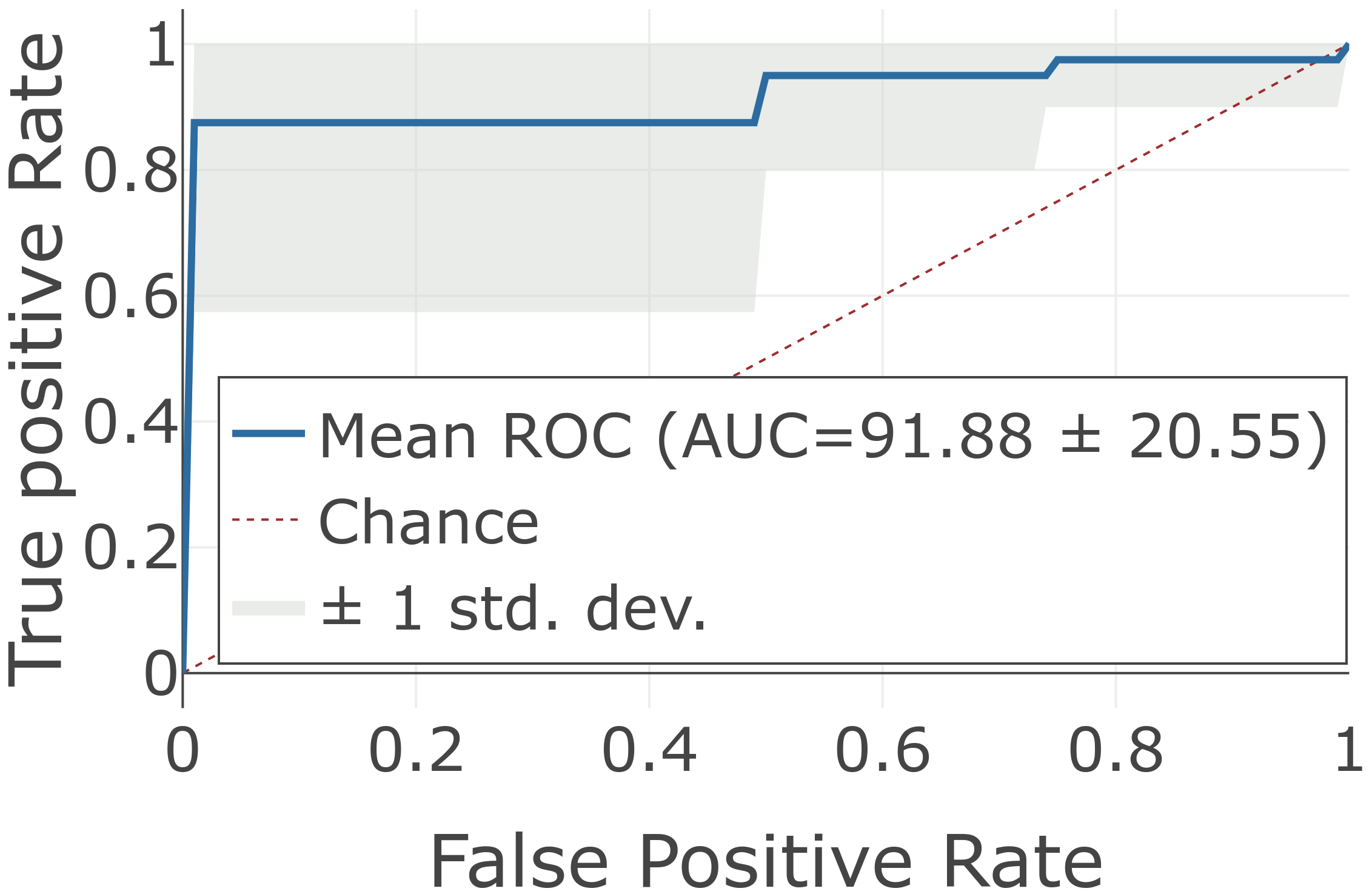}  
  \caption{ROC plot for Task 4}
\end{subfigure}
\\
\begin{subfigure}{.5\textwidth}
  \centering
  \includegraphics[width=.85\linewidth]{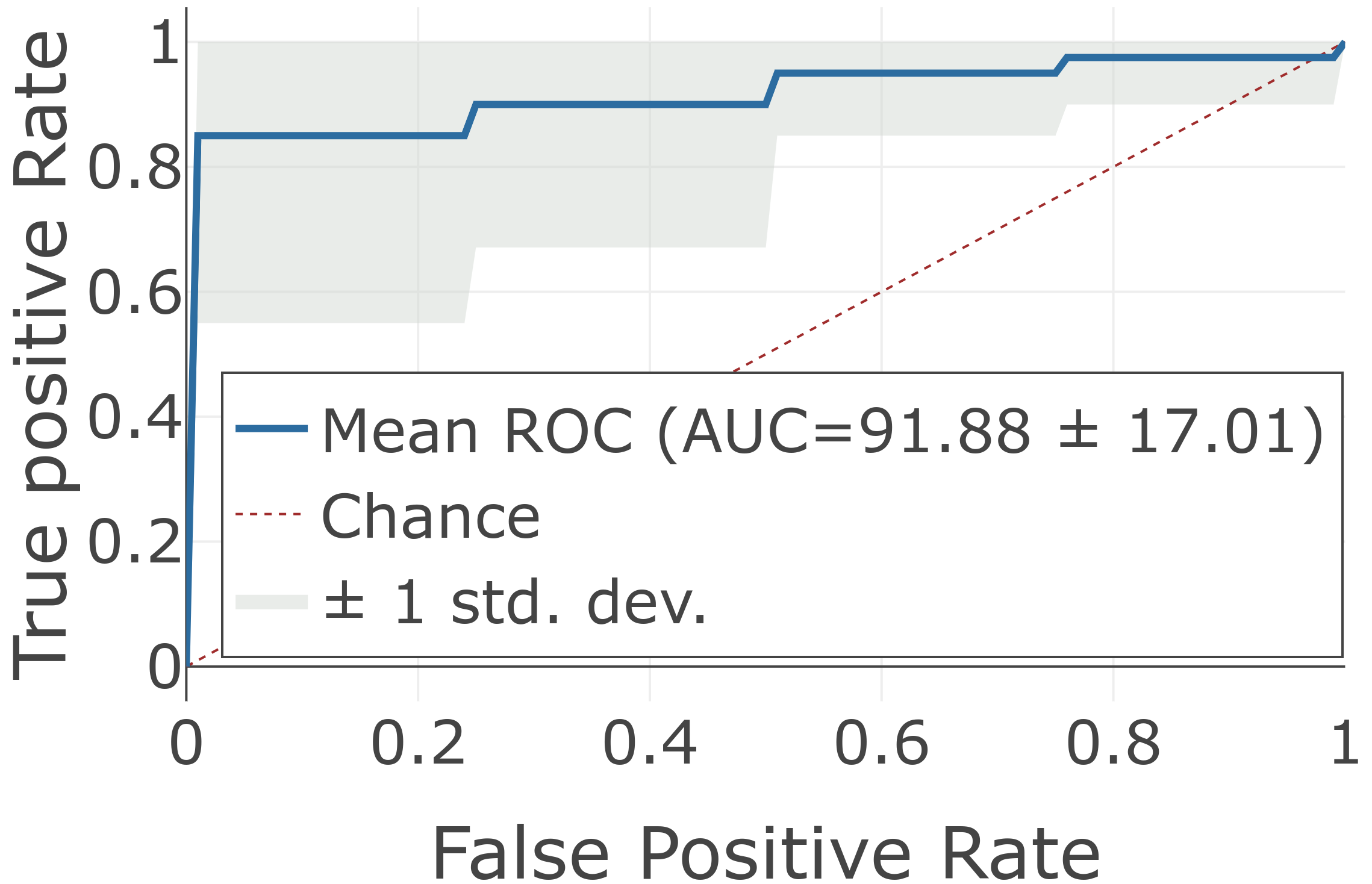}  
  \caption{ROC plot for Task 5}
\end{subfigure}
~
\begin{subfigure}{.5\textwidth}
  \centering
  \includegraphics[width=.85\linewidth]{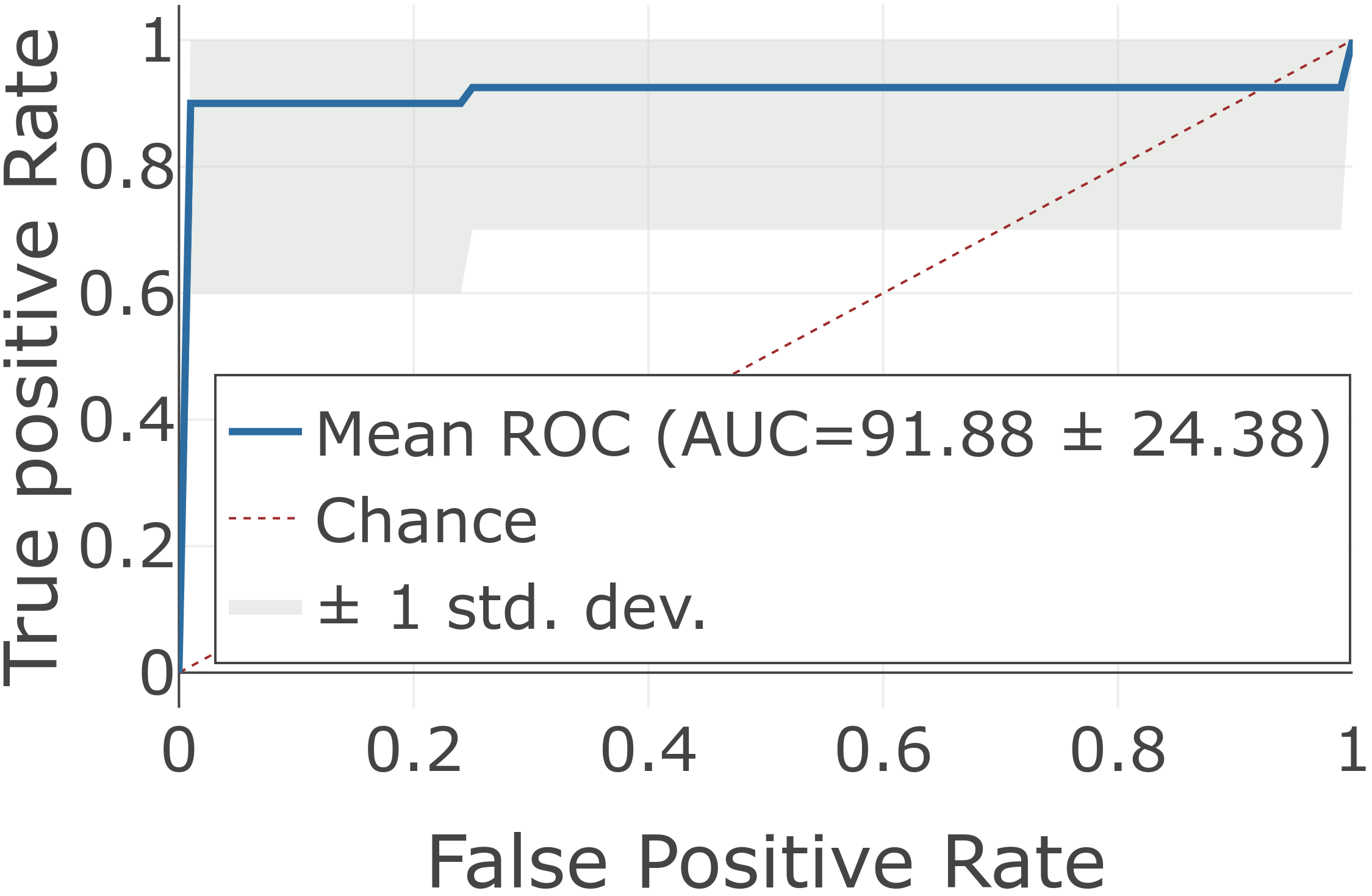}  
  \caption{ROC plot for Task 6}
\end{subfigure}
\\
\begin{subfigure}{.5\textwidth}
  \centering
  \includegraphics[width=.85\linewidth]{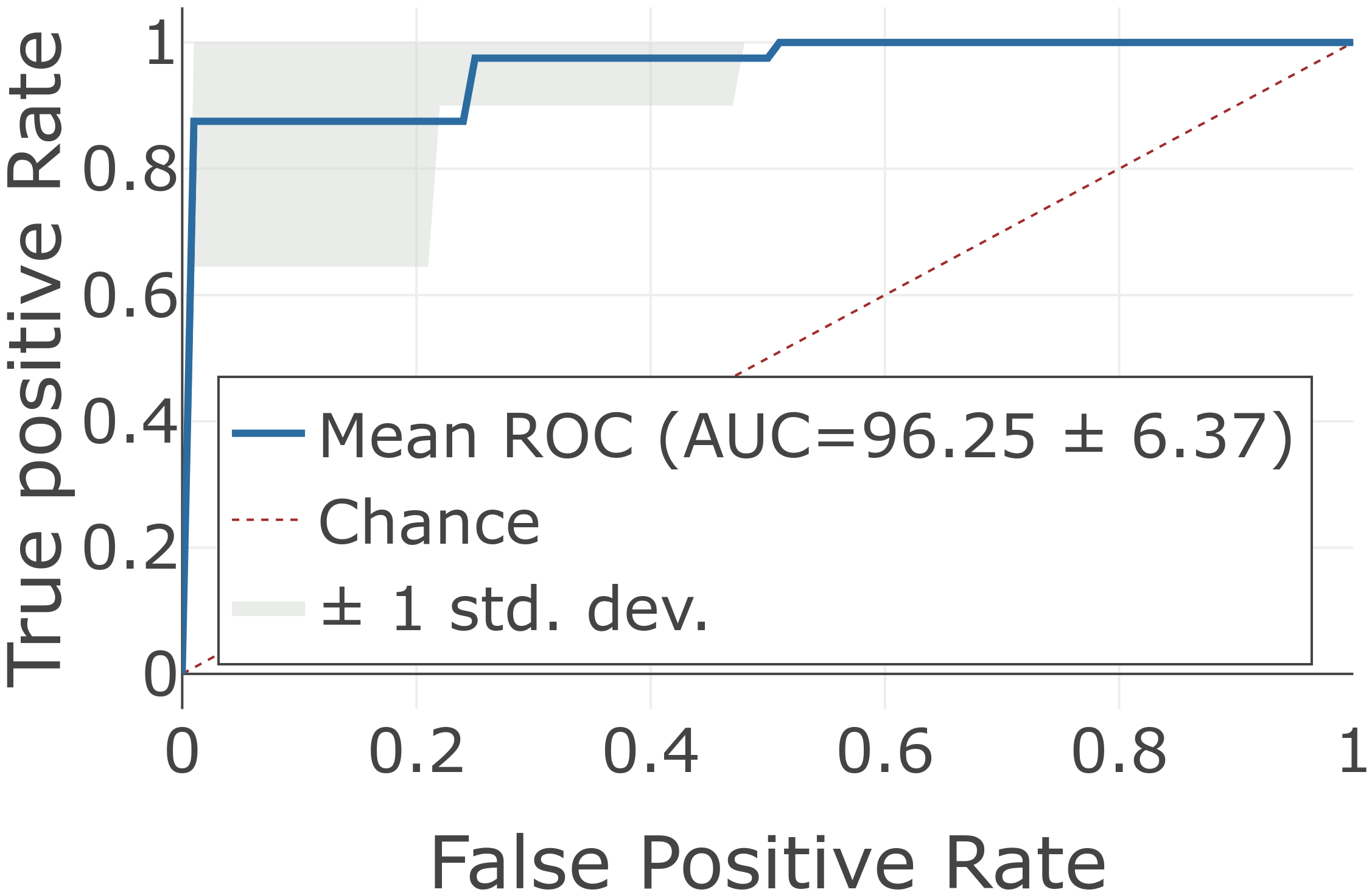}  
  \caption{ROC plot for Task 7}
\end{subfigure}
~
\begin{subfigure}{.5\textwidth}
  \centering
  \includegraphics[width=.85\linewidth]{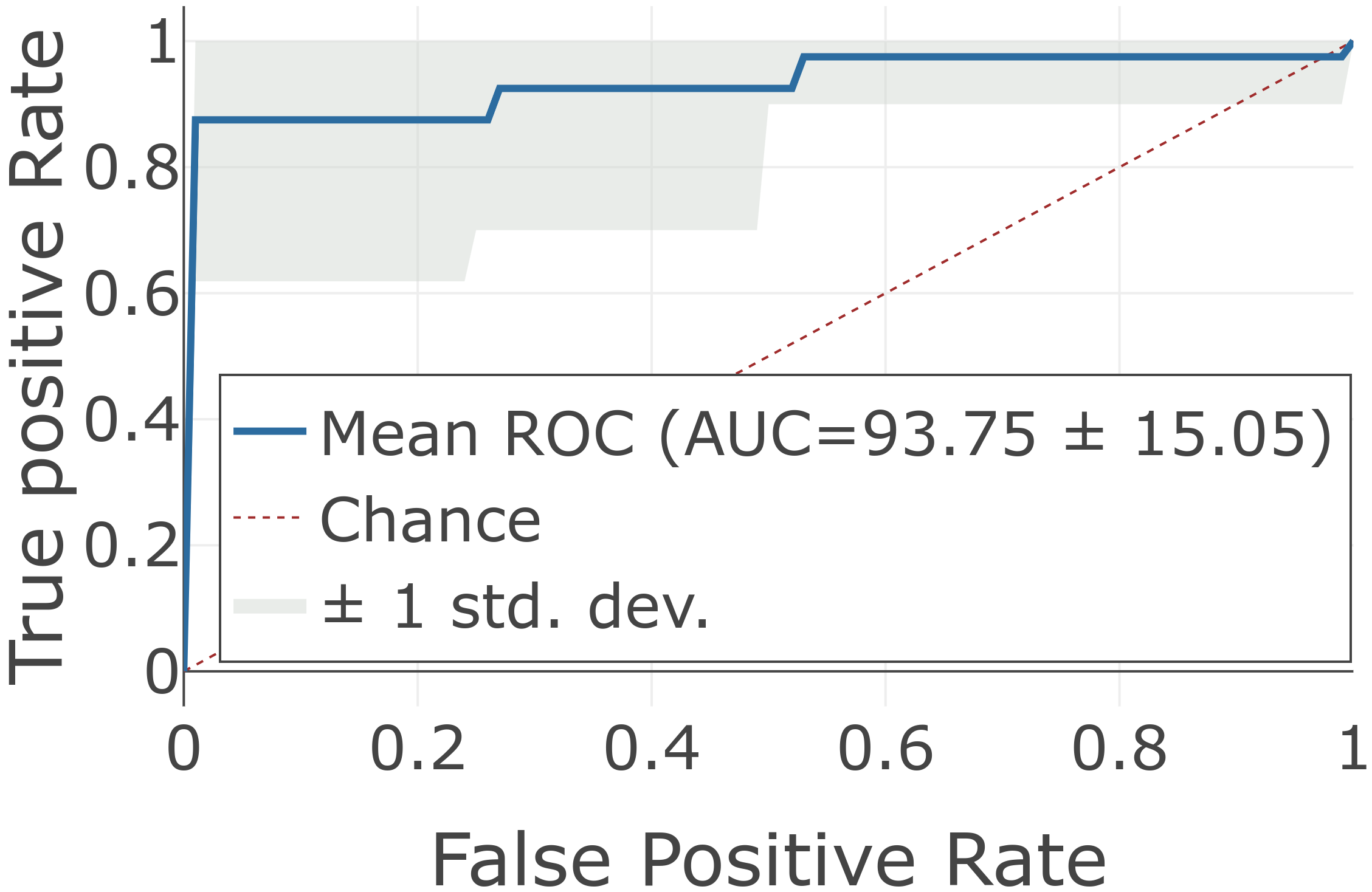}  
  \caption{ROC plot for Task 8}
\end{subfigure}
\caption{Task-wise ROC plots for the best feature set.}
\label{fig:roc_curves}
\end{figure}

In Table~\ref{tab:other_metrics}, we report the classification performance of the best performing feature set, i.e.~derived features, in terms of AUC, sensitivity, and specificity. In general, high values are obtained for all metrics and all tasks, confirming the applicability of the proposed method. There is usually a trade-off between sensitivity and specificity. In the present work, the method appears to be slightly biased in favor of specificity. This suggests that a screening test based on our tool will be better at correctly classifying healthy subjects. To further validate the proposed method, we also illustrate (in Fig.~\ref{fig:roc_curves}) the ROC plots for each of the eight tasks where the highest true positive and lowest false positive rates validate the robust discriminating power of the proposed model.

\begin{table}[!t]
\scriptsize
\centering
\begin{tabular}{l c c c c}
\hline
\textbf{Task} & \textbf{AUC} & \textbf{Sensitivity} & \textbf{Specificity}\\
\hline
Spiral & 93.12\% & 95.00\% & 92.50\% \\
\textit{lll} & 96.88\% & 92.50\% & 100.00\% \\
\textit{le le le} & 92.50\% & 85.00\% & 92.50\% \\
\textit{les les les} & 91.88\% & 92.50\% & 87.50\% \\
\textit{lektorka} & 91.88\% & 92.50\% & 95.00\% \\
\textit{porovnat} & 91.88\% & 87.50\% & 95.00\% \\
\textit{nepopadnout} & 96.25\% & 87.50\% & 97.50\% \\
Sentence & 93.75\% & 90.00\% & 95.00\% \\
\hline
\end{tabular}
\caption{Classification performance of the best performing feature set (derived features) in terms of other well-known metrics.}
\label{tab:other_metrics}
\end{table}

\subsection{Comparison with State-of-the-Art on PaHaW}
To further establish the effectiveness of the proposed method, a comparative analysis with state-of-the-art approaches aimed at detecting PD through the dynamics of handwriting is presented in Table~\ref{tab:soa}. It should be noted that, during the comparison, it was ensured that the same evaluation metric (mean accuracy) and the same validation scheme (10-fold cross-validation) were used. Indeed, this is the evaluation protocol usually adopted for PaHaW. Furthermore, since different classifiers (support vector machines, random forests, etc.) were used in these studies, reporting different classification accuracies, for a fair comparison, we report here only the best results of these studies for each task.\\
\begin{table}[!t]
\tiny
\centering
\begin{tabular}{lccccc}
\hline
    \textbf{Task} & \textbf{\citep{drotar2016evaluation}} & \textbf{\citep{impedovo2019velocity}} & \textbf{\citep{angelillo2019performance}} & \textbf{\citep{diaz2019dynamically}} & \textbf{\textit{This work}}\\
\hline
Spiral & 62.80\% & 97.33\% & 53.75\% & 75.00\% & 93.75\% \\
\textit{lll} & 72.30\% & 97.47\% & 67.08\% & 64.16\% & 96.25\% \\
\textit{le le le} & 71.00\% & 95.12\% & 72.50\% & 58.33\% & 88.75\% \\
\textit{les les les} & 66.40\% & 93.17\% & 57.91\% & 71.67\% & 90.00\% \\
\textit{lektorka} & 65.20\% & 96.79\% & 54.58\% & 75.41\% & 93.75\% \\
\textit{porovnat} & 73.30\% & 95.96\% & 63.75\% & 63.75\% & 91.25\% \\
\textit{nepopadnout} & 67.60\% & 96.76\% & 61.67\% & 70.00\% & 92.50\% \\
Sentence & 76.50\% & 92.05\% & 70.42\% & 67.08\% & 92.50\% \\
\hline
\end{tabular}
\caption{Performance comparison with state-of-the-art approaches on PaHaW.}
\label{tab:soa}
\end{table}

The first baseline selected for comparison consists of the traditional, hand-crafted dynamic features proposed by \citet{drotar2016evaluation}. In that work, the horizontal and vertical components of the pen position were segmented into on-surface and in-air strokes as a function of the button status value. Based on this segmentation, kinematic (displacement, velocity, acceleration, and jerk), spatio-temporal (on-surface time) and pressure features were derived. 
This feature extraction step resulted in either a single-valued feature or a vector feature. For all the resulting vector features, the following basic statistical measures were calculated: mean; median; standard deviation; 1st percentile; 99th percentile; 99th -- 1st percentile. 
The overall feature vectors were then fed into traditional statistical classifiers, after keeping the most discriminating subset with an \textit{a priori} selection of features made on the overall dataset before cross-validation. It is worth noting that supervised feature selection strategies should be \textit{nested} within the cross-validation iterations, so that the most discriminating features are chosen based only on the training set, while the test set is kept aside. Resorting to an \textit{a priori} selection of features over the entire dataset accidentally introduces a bias in the classification process which may lead to overoptimistic performance. This is well-known in the machine learning community; see, for example, \citep{hastie2009elements}.\\

The second baseline against which we compare the proposed method is the extension of the previously mentioned features proposed by \citet{impedovo2019velocity}. The author has enriched the conventional feature set with additional features obtained by applying the Sigma-Lognormal model, the Maxwell-Boltzmann distribution and the well-known Discrete Fourier Transform to the velocity profile of handwriting. Based on these features, significantly better results were obtained. However, it is worth pointing out that the author used the same \textit{non-nested} validation scheme previously applied by Drot\'ar {\it et al.} In fact, the author moved from their result as a baseline, then improved it. This means that, while remarkable, the provided contribution still suffers the same feature selection bias as the study of \citet{drotar2014decision}.\\

The third baseline consists of the results we obtained by replicating the experiments carried out by~\citet{drotar2014decision}, in which, in order to uncover hidden complexities of handwriting, features based on Shannon and R\'enyi entropy, signal-to-noise ratio, and empirical mode decomposition were also computed~\citep{angelillo2019performance}. In the work cited, we used a \textit{nested} feature selection in which the most discriminating features were chosen based only on the training set at each cross-validation iteration. In the work, we showed the detrimental effect on classification accuracy caused by inadvertently introducing the feature selection bias into the machine learning workflow.\\

Finally, the fourth baseline against which we compared the proposed method consists in the use of features automatically extracted by a 2D Convolutional Neural Network model~\citep{diaz2019dynamically}. More specifically, the well-known VGG16 deep network~\citep{simonyan2014very}, pre-trained on ImageNet~\citep{deng2009imagenet}, was applied as a feature extractor to multiple ``views'' of the same static representation of the handwriting. This representation was obtained by embedding dynamic temporal information during the image generation to retain velocity/in-air information. The overall feature vectors were fed into standard statistical classifiers to achieve the final classification, after dynamically retaining features with a \textit{nested} cross-validation scheme. \\

As can be inferred from Table~\ref{tab:soa}, the sequence learning approach based on 1D convolutions and BiGRUs significantly surpasses, by a considerable margin, the previously proposed procedures based on traditional dynamic features and 2D convolutions on the same dataset. This confirms the effectiveness of the proposed method to serve as a candidate solution for real use in a clinical setting. It is also worth noting that the feature selection bias problem does not apply here. Interestingly, all related works generally agree that the sentence task is among the most discriminating. Conversely, as noted earlier, the spiral task generally reports relatively lower performance when traditional dynamic features are used.

\subsection{Ablation Study}
We also report the results of ablation studies we carried out to choose the model architecture. These results are summarized in Table~\ref{tab:first_ablation} and Table~\ref{tab:second_ablation}. It is worth mentioning that these results were obtained by feeding the models with the derived feature set. The first ablation study (Table~\ref{tab:first_ablation}) was conducted to observe the performance of different RNN-based models, which served as a baseline for the outcome of the second ablation study. The features were fed directly into the recurrent units without any convolutional layers. It was observed that the Bidirectional GRU-based model outperformed the other RNN variants on each of the tasks.\\

The second ablation study (Table~\ref{tab:second_ablation}) mainly concerned the evaluation of the effectiveness of jointly exploiting convolutional and recurrent units. It can be observed that applying convolution on the sequences provided as input, before feeding them into the BiGRU layers, significantly improves classification accuracy. Moreover, it is worth pointing out that, due to the sub-sampling of the time sequences obtained using different stride size (greater than one), the training complexity of all RNN models is also reduced. Overall, these findings further validate our hypothesis that time-based handwriting sequences contain unique patterns that can be enhanced by convolution and identified by Bidirectional GRUs for PD classification. 

\begin{table}[!t]
\scriptsize
\centering
\begin{tabular}{l c c c c}
\hline
\textbf{Task} & \textbf{BiRNN} & \textbf{BiLSTM} & \textbf{BiGRU}\\
\hline
Spiral & 84.29\% & 87.86\% & 88.57\% \\
\textit{lll} & 81.07\% & 83.39\% & 83.57\% \\
\textit{le le le} & 75.71\% & 75.71\% & 82.32\% \\
\textit{les les les} & 79.64\% & 80.00\% & 84.82\% \\
\textit{lektorka} & 74.11\% & 75.89\% & 80.00\% \\
\textit{porovnat} & 77.14\% & 78.39\% & 82.32\% \\
\textit{nepopadnout} & 75.54\% & 82.32\% & 83.75\% \\
Sentence & 83.57\% & 85.00\% & 86.25\% \\
\hline
\end{tabular}
\caption{Comparison between different RNN models without convolution.}
\label{tab:first_ablation}
\end{table}

\begin{table}[!t]
\scriptsize
\centering
\begin{tabular}{l c c c c}
\hline
\textbf{Task} & \textbf{BiRNN} & \textbf{BiLSTM} & \textbf{BiGRU}\\
\hline
Spiral & 88.33\% & 90.00\% & 93.75\% \\
\textit{lll} & 91.25\% & 94.38\% & 96.25\% \\
\textit{le le le} & 85.00\% & 88.50\% & 88.75\% \\
\textit{les les les} & 87.50\% & 89.67\% & 90.00\% \\
\textit{lektorka} & 88.75\% & 92.38\% & 93.75\% \\
\textit{porovnat} & 87.50\% & 88.75\% & 91.25\% \\
\textit{nepopadnout} & 89.40\% & 91.00\% & 92.50\% \\
Sentence & 90.00\% & 92.32\% & 92.50\% \\
\hline
\end{tabular}
\caption{Comparison between different RNN models with 1D convolution.}
\label{tab:second_ablation}
\end{table}

\subsection{Classification Results on NewHandPD}

To validate the generalization capacity of our approach, we ran a series of experiments on NewHandPD. This dataset was not seen during the configuration of our system. Therefore, the best configuration found has been used here.\\

Specifically, the experiments with NewHandPD have been performed using an experimental protocol similar to that used in~\citep{ribeiro2019bag}. This included using 65\% of the data for training, 10\% for validation and 25\% for testing. 
The results obtained have been averaged after repeating the experiment for 20 runs. It is worth noting that we applied the same data preprocessing suggested in \citep{ribeiro2019bag}, which consisted of removing outliers by cutting off values below the 5th percentile and above the 90th percentile on each channel. Moreover, a $z$-score normalization was applied.\\

The experimental results are provided in Table~\ref{tab:other_metrics2} for each type of exam. Similar to previous results, we reported performance in terms of AUC, sensitivity, specificity and accuracy. Competitive results have been obtained, especially for the spiral-based exam. In addition, the specificity results are consistently greater than sensitivity in all cases. These two effects give consistency to our method as performance with PaWaH also showed similar findings.\\

\begin{table}[!t]
\scriptsize
\centering
\begin{tabular}{l c c c c c }
\hline
\textbf{Task} & \textbf{AUC} & \textbf{Sensitivity} & \textbf{Specificity} & \textbf{Accuracy}\\
\hline
Spiral & 98.25\%   & 90.00\%   &98.00\%  & 94.44\%\\ 
Meander & 97.75\% &90.00\% & 92.00\% & 91.11\% \\
Circle$_\text{s}^\dag$ & 92.25\% & 85.00\% & 92.00\% & 88.89\% \\
Circle$_\text{a}^\dag$ & 85.91\% & 85.62\% & 85.50\% & 85.56\% \\
Diadochokinesis$_\text{R}$ & 71.00\% & 55.00\% & 78.00\% & 67.78\% \\
Diadochokinesis$_\text{L}$ & 73.50\% & 65.00\% & 76.00\% & 71.11\% \\\hline
\multicolumn{5}{l}{Circled movements on surface and in the air.}\\
\end{tabular}
\caption{Classification performance on NewHandPD following the experimental protocol of~\citep{ribeiro2019bag} with our method.}
\label{tab:other_metrics2}
\end{table}

Furthermore, we have analyzed the previous literature with this particular database in order to contextualize our results. In~\citep{pereira2016deep}, the authors proposed to model the time-series of the NewHandPD dataset as images. The images were designed using the six available channels as well as the temporal sequences. Then, different CNN architectures were exploited. Specifically, the authors experimented with spirals and meanders. Among the different configurations studied, their best results are shown in Table~\ref{tab:soa2}. They correspond to a network pre-trained on ImageNet, accepting $128 \times 128$ images and using 75\,\% of the dataset for training. The authors also reported the results obtained using other classifiers, such as Optimum-Path Forest.\\

\begin{table}[!t]
\scriptsize
\centering
\begin{tabular}{lcccc}
\hline
    \textbf{Task} & \textbf{(Pereira et al.)}& \textbf{(Pereira et al.)} & \textbf{(Ribeiro et al.)} & \textbf{\textit{This work}}\\
     & \textbf{(2016)} & \textbf{(2018)} & \textbf{(2019)} & \\
\hline
Spiral &    77.53\% & 78.26\% &89.48\%  &94.44\%\\
Meanders & 87.14\% &80.75\% & 92.24\% & 91.11\% \\
Circle$_\text{s}^\dag$ & - & 68.04\% & - & 88.89\% \\
Circle$_\text{a}^\dag$ & - & 73.41\% & - & 85.56\% \\
Diadochokinesis$_\text{R}$ & - & 73.59\% & - & 67.78\% \\
Diadochokinesis$_\text{L}$ & - & 76.32\% & - & 71.11\% \\\hline
\multicolumn{5}{l}{Circled movements on surface and in the air.}\\
\multicolumn{5}{l}{``-'' means results not reported.}\\
\end{tabular}
\caption{Performance comparison with state-of-the-art approaches on NewHandPD.}
\label{tab:soa2}
\end{table}

A similar approach was presented in~\citep{pereira2018handwritten}. Again, several CNN architectures were investigated to discriminate between healthy controls and PD patients. In this case, all exams available in NewHandPD were studied using samples from 20 healthy controls and 14 PD patients. The results shown in Table~\ref{tab:soa2} were achieved with 50\,\% of the specimens for training and using the ImageNet-based network. As a step forward, the authors also presented the results obtained by combining all the exams to achieve a single classification.\\

For the most relevant comparison, we have selected the work of~\citep{ribeiro2019bag}, in which stacks of Bidirectional Gated Recurrent Units were employed with an attention layer on top. The authors introduced a bag-of-sampling concept for selecting samples of signal sequences provided in the NewHandPD dataset. The results show that this approach led to better classification outcomes compared to previous studies (Table~\ref{tab:soa2}). The experiments were performed using 65\% of data for training, 10\% for validation and 25\% for testing. Instead of using the entire database, the authors used 25 control subjects and 14 patients.\\

The overall results of our proposed model on the NewHandPD data with a similar experimental protocol show a significant improvement in classification when compared to the results of~\cite{pereira2016deep} and~\cite{pereira2018handwritten}. When compared with~\cite{ribeiro2019bag}, it is observed that our method has improved results when the spiral is used for classification, while in the case of meanders, our method behaves comparatively with~\citep{ribeiro2019bag}. It is also noteworthy that our results are computed using all the samples, i.e.~35 healthy and 31 diseased subjects, provided in the NewHandPD database.\\

Finally, it is worth pointing out that all the results shown correspond to the best configurations studied in each paper for the NewHandPD database. In our work, we have used this database only to demonstrate the generalization capacity of our approach.

\section{Conclusion}
\label{sec:c}
The growing body of evidence on computerized dynamic handwriting analysis supports the hypothesis that handwriting measures can capture the physical and cognitive characteristics of individuals. In particular, since handwriting difficulties in Parkinsonian patients have been documented for a long time, such an analysis is promising to help assess Parkinson's disease. The best prospect of this line of research is the integration of new medical tools into current clinical practices to increase the level of diagnostic accuracy. Domain experts can be provided with these easy-to-use, user-friendly tools in their daily practice, without the need for any specific computing expertise. In this sense, a handwriting-based tool represents an attractive choice as it not only provides professionals with a prompt automatic response, but also allows them to store useful metadata related to patient medical records for later use. Of course, handwriting-based decision support tools are not expected to replace standard techniques or humans, but rather provide additional evidence to support their clinical assessment.\\

In this study, we have proposed a new model based on one-dimensional convolutions and Bidirectional GRUs to identify distinctive patterns in the handwriting sequences of PD patients and controls. Different sets of dynamic features acquired from on-line graphomotor samples of both groups were fed to the model as input. Convolutional layers perform sub-sampling and learn effective feature representations before sending sequences to the Bidirectional GRU part of the network. The results of our experimental study indicate the effectiveness of the proposed technique with respect to the state-of-the-art. The proposed method, in fact, outperformed other ``holistic'' approaches, thus confirming the effectiveness of the sequence learning paradigm for processing sequential handwriting data. We believe that in addition to the quantitative results, providing a new perspective on the same problem can help clarify some underlying mechanisms still unknown in the future and offer new insights that may be particularly useful for this specific domain. Another observation concerns the exploitation of two datasets whose specimens have been acquired through different technologies. Although previous and recent literature typically used a single dataset for model development and evaluation, the use of a second dataset helped us confirm the robustness of the proposed method.\\

A significant limitation of the present study is the small size of the datasets we employed, which can somewhat influence the generalizability of the results obtained. Unfortunately, developing a large benchmark dataset is still one of the major open issues in the pattern recognition community working in this field~\citep{vessio2019dynamic,2021_COGN_HandwTrends_Faundez}. This applies not only to PD, but also to other neurodegenerative disorders~\citep{impedovo2019handwriting}. Nevertheless, despite these constraints, the reported performance values are indeed very promising and the results of this study are expected to make way for a working system in the clinical settings.

\section*{Authorship Contribution Statement}
{\bf Moises~Diaz:} Conceptualization, Validation, Writing - Original Draft, Writing - Review \& Editing, Visualization. {\bf Momina~Moetesum:} Methodology, Software, Investigation, Resources, Writing - Review \& Editing. {\bf Imran~Siddiqi:} Conceptualization, Investigation, Writing - Review \& Editing, Supervision. {\bf Gennaro~Vessio:}  Conceptualization, Validation, Writing - Original Draft, Writing - Review \& Editing.
 
\section*{Declaration of Interest}
The authors declare that they have no known competing financial interests or personal relationships that could have appeared to influence the work reported in this paper.

\section*{Acknowledgment}
Part of this research was funded by the Higher Education Commission (HEC), Pakistan, under grant number 8910/Federal/NRPU/R\&D/HEC/2017. Additional funding was received from the Spanish government's MIMECO TEC2016-77791-C4-1-R and PID2019-109099RB-C41 research projects and the European Union FEDER program/funds.






\bibliographystyle{agsm}
\biboptions{authoryear}
\bibliography{main.bib}

@article{pereira2016new,
  title={A new computer vision-based approach to aid the diagnosis of Parkinson's disease},
  author={Pereira, Clayton R and Pereira, Danilo R and Silva, Francisco A and Masieiro, Jo{\~a}o P and Weber, Silke AT and Hook, Christian and Papa, Jo{\~a}o P},
  journal={Computer Methods and Programs in Biomedicine},
  volume={136},
  pages={79--88},
  year={2016},
  publisher={Elsevier}
}

@inproceedings{pereira2016deep,
  title={Deep learning-aided Parkinson's disease diagnosis from handwritten dynamics},
  author={Pereira, Clayton and Weber, Silke  and Hook, Christian and Rosa, Gustavo  and Papa, Joao },
  booktitle={2016 29th SIBGRAPI Conference on Graphics, Patterns and Images (SIBGRAPI)},
  pages={340--346},
  year={2016},
  organization={Ieee}
}

@article{ferrer2020idelog,
  title={i{D}e{L}og: iterative dual spatial and kinematic extraction of sigma-lognormal parameters},
  author={Ferrer, Miguel A and Diaz, Moises and Carmona-Duarte, Cristina and Plamondon, R{\'e}jean},
  journal={IEEE Transactions on Pattern Analysis and Machine Intelligence},
  volume={42},
  number={1},
  pages={114--125},
  year={2020},
  publisher={IEEE},
  note={\url{https://doi.org/10.1109/TPAMI.2018.2879312}}
}

@incollection{2020_LognormalityChapter_Diaz,
author = {Moises Diaz and  Miguel A. Ferrer and Cristina Carmona and Réjean Plamondon},
booktitle = {The Lognormality Principle and its Applications},
editor = {Rejean Plamondon and Angelo Marcelli and Miguel A. Ferrer },
publisher = {World Scientific},
title = {Improving Handwritten Signatures Fluency via the Lognormality Principles},
year = {2021},
pages={41-63},
note={\url{https://doi.org/10.1142/9789811226830 0002}}
}

@ARTICLE{2021_COGN_HandwTrends_Faundez,
  title={Handwriting biometrics: Applications and future trends in e-security and e-health},
  author={Faundez-Zanuy, Marcos and Fierrez, Julian and Ferrer, Miguel A and Diaz, Moises and Tolosana, Ruben and Plamondon, R{\'e}jean},
  journal={Cognitive Computation},
  volume={12},
  number={5},
  pages={940--953},
  year={2020},
  publisher={Springer},
note={\url{https://doi.org/10.1007/s12559-020-09755-z}},
}

@article{diaz2019perspective,
  title={A perspective analysis of handwritten signature technology},
  author={Diaz, Moises and others},
  journal={ACM Computing Surveys (CSUR)},
  volume={51},
  number={6},
  pages={1--39},
  year={2019},
  publisher={ACM New York, NY, USA},
  note={\url{https://doi.org/10.1145/3274658}}
}

@article{ascherio2016epidemiology,
  title={The epidemiology of {P}arkinson's disease: risk factors and prevention},
  author={Ascherio, Alberto and Schwarzschild, Michael A},
  journal={The Lancet Neurology},
  volume={15},
  number={12},
  pages={1257--1272},
  year={2016},
  publisher={Elsevier},
  note={\url{https://doi.org/10.1016/S1474-4422(16)30230-7}},
}

@article{vessio2019dynamic,
  title={Dynamic Handwriting Analysis for Neurodegenerative Disease Assessment: A Literary Review},
  author={Vessio, Gennaro},
  journal={Applied Sciences},
  volume={9},
  number={21},
  pages={4666},
  year={2019},
  publisher={Multidisciplinary Digital Publishing Institute},
  note={\url{https://doi.org/10.3390/app9214666}}
}

@article{rosenblum2013handwriting,
  title={Handwriting as an objective tool for {P}arkinson's disease diagnosis},
  author={Rosenblum, Sara and Samuel, Margalit and Zlotnik, Sharon and Erikh, Ilana and Schlesinger, Ilana},
  journal={Journal of Neurology},
  volume={260},
  number={9},
  pages={2357--2361},
  year={2013},
  publisher={Springer},
  note={\url{https://doi.org/10.1007/s00415-013-6996-x}}
}

@article{drotar2016evaluation,
  title={Evaluation of handwriting kinematics and pressure for differential diagnosis of {P}arkinson's disease},
  author={Drot{\'a}r, Peter and Mekyska, Ji{\v{r}}{\'\i} and Rektorov{\'a}, Irena and Masarov{\'a}, Lucia and Sm{\'e}kal, Zden{\v{e}}k and Faundez-Zanuy, Marcos},
  journal={Artificial Intelligence in Medicine},
  volume={67},
  pages={39--46},
  year={2016},
  publisher={Elsevier},
  note={\url{https://doi.org/10.1016/j.artmed.2016.01.004}}
}

@article{ammour2020new,
  title={A new semi-supervised approach for characterizing the Arabic on-line handwriting of {P}arkinson's disease patients},
  author={Ammour, Alae and Aouraghe, Ibtissame and Khaissidi, Ghizlane and Mrabti, Mostafa and Aboulem, Ghita and Belahsen, Faouzi},
  journal={Computer Methods and Programs in Biomedicine},
  volume={183},
  pages={104979},
  year={2020},
  publisher={Elsevier},
  note={\url{https://doi.org/10.1016/j.cmpb.2019.07.007}},
}

@article{drotar2014analysis,
  title={Analysis of in-air movement in handwriting: A novel marker for {P}arkinson's disease},
  author={Drot{\'a}r, Peter and Mekyska, Ji{\v{r}}{\'\i} and Rektorov{\'a}, Irena and Masarov{\'a}, Lucia and Sm{\'e}kal, Zdenek and Faundez-Zanuy, Marcos},
  journal={Computer Methods and Programs in Biomedicine},
  volume={117},
  number={3},
  pages={405--411},
  year={2014},
  publisher={Elsevier},
  note={\url{https://doi.org/10.1016/j.cmpb.2014.08.007}}
}

@article{drotar2014decision,
  title={Decision support framework for {P}arkinson's disease based on novel handwriting markers},
  author={Drot{\'a}r, Peter and Mekyska, Ji{\v{r}}{\'\i} and Rektorov{\'a}, Irena and Masarov{\'a}, Lucia and Sm{\'e}kal, Zden{\v{e}}k and Faundez-Zanuy, Marcos},
  journal={IEEE Transactions on Neural Systems and Rehabilitation Engineering},
  volume={23},
  number={3},
  pages={508--516},
  year={2015},
  publisher={IEEE},
  note={\url{https://doi.org/10.1109/TNSRE.2014.2359997}}
}

@article{impedovo2019velocity,
  title={Velocity-based signal features for the assessment of {P}arkinsonian handwriting},
  author={Impedovo, Donato},
  journal={IEEE Signal Processing Letters},
  volume={26},
  number={4},
  pages={632--636},
  year={2019},
  publisher={IEEE},
  note={\url{https://doi.org/10.1109/LSP.2019.2902936}}
}

@article{diaz2019dynamically,
  title={Dynamically enhanced static handwriting representation for {P}arkinson's disease detection},
  author={Diaz, Moises and others},
  journal={Pattern Recognition Letters},
  volume={128},
  pages={204--210},
  year={2019},
  publisher={Elsevier},
  note={\url{https://doi.org/10.1016/j.patrec.2019.08.018}}
}

@article{pereira2018handwritten,
  title={Handwritten dynamics assessment through convolutional neural networks: An application to {P}arkinson's disease identification},
  author={Pereira, Clayton R and Pereira, Danilo R and Rosa, Gustavo H and Albuquerque, Victor HC and Weber, Silke AT and Hook, Christian and Papa, Jo{\~a}o P},
  journal={Artificial Intelligence in Medicine},
  volume={87},
  pages={67--77},
  year={2018},
  publisher={Elsevier},
  note={\url{https://doi.org/10.1016/j.artmed.2018.04.001}}
}

@article{de2019handwriting,
  title={Handwriting analysis to support neurodegenerative diseases diagnosis: A review},
  author={De Stefano, Claudio and Fontanella, Francesco and Impedovo, Donato and Pirlo, Giuseppe and di Freca, Alessandra Scotto},
  journal={Pattern Recognition Letters},
  volume={121},
  pages={37--45},
  year={2019},
  publisher={Elsevier},
  note={\url{https://doi.org/10.1016/j.patrec.2018.05.013}}
}

@article{hochreiter1997long,
  title={Long short-term memory},
  author={Hochreiter, Sepp and Schmidhuber, J{\"u}rgen},
  journal={Neural Computation},
  volume={9},
  number={8},
  pages={1735--1780},
  year={1997},
  publisher={MIT Press},
  note={\url{https://doi.org/10.1162/neco.1997.9.8.1735}}
}

@inproceedings{cho2014learning,
  title={Learning phrase representations using {RNN} encoder-decoder for statistical machine translation},
  author={Cho, Kyunghyun and Van Merri{\"e}nboer, Bart and Gulcehre, Caglar and Bahdanau, Dzmitry and Bougares, Fethi and Schwenk, Holger and Bengio, Yoshua},
  year={2014},
  note={\url{arXiv:1406.1078}},
}

@article{smits2014standardized,
  title={Standardized handwriting to assess bradykinesia, micrographia and tremor in {P}arkinson's disease},
  author={Smits, Esther J and Tolonen, Antti J and Cluitmans, Luc and Van Gils, Mark and Conway, Bernard A and Zietsma, Rutger C and Leenders, Klaus L and Maurits, Natasha M},
  journal={PloS one},
  volume={9},
  number={5},
  pages={e97614},
  year={2014},
  publisher={Public Library of Science},
  note={\url{https://doi.org/10.1371/journal.pone.0097614}}
}

@article{senatore2019paradigm,
  title={A paradigm for emulating the early learning stage of handwriting: Performance comparison between healthy controls and {P}arkinson's disease patients in drawing loop shapes},
  author={Senatore, Rosa and Marcelli, Angelo},
  journal={Human Movement Science},
  volume={65},
  pages={89--101},
  year={2019},
  publisher={Elsevier},
  note={\url{https://doi.org/10.1016/j.humov.2018.04.007}}
}

@article{randhawa2013repetitive,
  title={Repetitive transcranial magnetic stimulation improves handwriting in {P}arkinson's disease},
  author={Randhawa, Bubblepreet K and Farley, Becky G and Boyd, Lara A},
  journal={Parkinson's Disease},
  volume={2013},
  year={2013},
  publisher={Hindawi},
  note={\url{https://doi.org/10.1155/2013/751925}}
}

@article{danna2019digitalized,
  title={Digitalized spiral drawing in {P}arkinson's disease: A tool for evaluating beyond the written trace},
  author={Danna, J{\'e}r{\'e}my and Velay, Jean-Luc and Eusebio, Alexandre and V{\'e}ron-Delor, Lauriane and Witjas, Tatiana and Azulay, Jean-Philippe and Pinto, Serge},
  journal={Human Movement Science},
  volume={65},
  pages={80--88},
  year={2019},
  publisher={Elsevier},
  note={\url{https://doi.org/10.1016/j.humov.2018.08.003}}
}

@article{moetesum2019assessing,
  title={Assessing visual attributes of handwriting for prediction of neurological disorders--{A} case study on {P}arkinson's disease},
  author={Moetesum, Momina and Siddiqi, Imran and Vincent, Nicole and Cloppet, Florence},
  journal={Pattern Recognition Letters},
  volume={121},
  pages={19--27},
  year={2019},
  publisher={Elsevier},
  note={\url{https://doi.org/10.1016/j.patrec.2018.04.008}}
}

@article{rios2019analysis,
  title={Analysis and evaluation of handwriting in patients with {P}arkinson's disease using kinematic, geometrical, and non-linear features},
  author={Rios-Urrego, Cristian D and V{\'a}squez-Correa, Juan Camilo and Vargas-Bonilla, Jes{\'u}s Francisco and N{\"o}th, Elmar and Lopera, Francisco and Orozco-Arroyave, Juan Rafael},
  journal={Computer Methods and Programs in Biomedicine},
  volume={173},
  pages={43--52},
  year={2019},
  publisher={Elsevier},
  note={\url{https://doi.org/10.1016/j.cmpb.2019.03.005}}
}

@inproceedings{Moetesum2020dynamic,
  title={Dynamic Handwriting Analysis for {P}arkinson's Disease Identification using {C}-{B}i{GRU} Model},
  author={Moetesum, Momina and Siddiqi, Imran and Javed, Farah and Masroor, Uzma},
  booktitle={2020 17th International Conference on Frontiers in Handwriting Recognition (ICFHR)},
  pages={115--120},
  year={2020},
  organization={IEEE}
}

@article{afonso2019recurrence,
  title={A recurrence plot-based approach for {P}arkinson's disease identification},
  author={Afonso, Luis CS and Rosa, Gustavo H and Pereira, Clayton R and Weber, Silke AT and Hook, Christian and Albuquerque, Victor Hugo C and Papa, Jo{\~a}o P},
  journal={Future Generation Computer Systems},
  volume={94},
  pages={282--292},
  year={2019},
  note={\url{https://doi.org/10.1016/j.future.2018.11.054}},
  publisher={Elsevier},
}

@article{ribeiro2019bag,
  title={Bag of Samplings for computer-assisted {P}arkinson's disease diagnosis based on Recurrent Neural Networks},
  author={Ribeiro, Luiz CF and Afonso, Luis CS and Papa, Jo{\~a}o P},
  journal={Computers in Biology and Medicine},
  volume={115},
  pages={103477},
  year={2019},
  publisher={Elsevier},
  note={\url{https://doi.org/10.1016/j.compbiomed.2019.103477}}
}

@inproceedings{angelillo2019performance,
  title={Performance-Driven Handwriting Task Selection for {P}arkinson's Disease Classification},
  author={Angelillo, Maria Teresa and Impedovo, Donato and Pirlo, Giuseppe and Vessio, Gennaro},
  booktitle={International Conference of the Italian Association for Artificial Intelligence},
  pages={281--293},
  year={2019},
  organization={Springer},
  note={\url{https://doi.org/10.1007/978-3-030-35166-3_20}},
}

@article{bhat2018parkinson,
  title={{P}arkinson's disease: Cause factors, measurable indicators, and early diagnosis},
  author={Bhat, Shreya and Acharya, U Rajendra and Hagiwara, Yuki and Dadmehr, Nahid and Adeli, Hojjat},
  journal={Computers in Biology and Medicine},
  volume={102},
  pages={234--241},
  year={2018},
  publisher={Elsevier},
  note={\url{https://doi.org/10.1016/j.compbiomed.2018.09.008}}
}

@article{parisi2018feature,
  title={Feature-driven machine learning to improve early diagnosis of {P}arkinson's disease},
  author={Parisi, Luca and RaviChandran, Narrendar and Manaog, Marianne Lyne},
  journal={Expert Systems with Applications},
  volume={110},
  pages={182--190},
  year={2018},
  publisher={Elsevier},
  note={\url{https://doi.org/10.1016/j.eswa.2018.06.003}}
}

@article{ali2019early,
  title={Early diagnosis of {P}arkinson's disease from multiple voice recordings by simultaneous sample and feature selection},
  author={Ali, Liaqat and Zhu, Ce and Zhou, Mingyi and Liu, Yipeng},
  journal={Expert Systems with Applications},
  volume={137},
  pages={22--28},
  year={2019},
  publisher={Elsevier},
  note={\url{https://doi.org/10.1016/j.eswa.2019.06.052}},
}

@article{broderick2009hypometria,
  title={Hypometria and bradykinesia during drawing movements in individuals with {P}arkinson's disease},
  author={Broderick, Michael P and Van Gemmert, Arend WA and Shill, Holly A and Stelmach, George E},
  journal={Experimental Brain Research},
  volume={197},
  number={3},
  pages={223--233},
  year={2009},
  publisher={Springer},
  note={\url{https://doi.org/10.1007/s00221-009-1925-z}}
}

@article{yu2019review,
  title={A review of recurrent neural networks: {LSTM} cells and network architectures},
  author={Yu, Yong and Si, Xiaosheng and Hu, Changhua and Zhang, Jianxun},
  journal={Neural Computation},
  volume={31},
  number={7},
  pages={1235--1270},
  year={2019},
  publisher={MIT Press},
  note={\url{https://doi.org/10.1162/neco_a_01199}}
}

@article{jerkovic2019analysis,
  title={Analysis of on-surface and in-air movement in handwriting of subjects with {P}arkinson's disease and atypical parkinsonism},
  author={Jerkovic, Vera Miler and Kojic, Vladimir and Miskovic, Natasa Dragasevic and Djukic, Tijana and Kostic, Vladimir S and Popovic, Mirjana B},
  journal={Biomedical Engineering/Biomedizinische Technik},
  volume={64},
  number={2},
  pages={187--194},
  year={2019},
  publisher={De Gruyter},
  note={\url{https://doi.org/10.1515/bmt-2017-0148}}
}

@article{kotsavasiloglou2017machine,
  title={Machine learning-based classification of simple drawing movements in {P}arkinson's disease},
  author={Kotsavasiloglou, C and Kostikis, N and Hristu-Varsakelis, Dimitrios and Arnaoutoglou, M},
  journal={Biomedical Signal Processing and Control},
  volume={31},
  pages={174--180},
  year={2017},
  publisher={Elsevier},
  note={\url{https://doi.org/10.1016/j.bspc.2016.08.003}}
}

@article{linden2018dynamic,
  title={Dynamic signatures: A review of dynamic feature variation and forensic methodology},
  author={Linden, Jacques and Marquis, Raymond and Bozza, Silvia and Taroni, Franco},
  journal={Forensic science international},
  volume={291},
  pages={216--229},
  year={2018},
  publisher={Elsevier},
  note={\url{https://doi.org/10.1016/j.forsciint.2018.08.021}}
}

@article{bidet2011handwriting,
  title={Handwriting in patients with {P}arkinson disease: effect of {L}-dopa and stimulation of the sub-thalamic nucleus on motor anticipation},
  author={Bidet-Ildei, Christel and Pollak, Pierre and Kandel, Sonia and Fraix, Val{\'e}rie and Orliaguet, Jean-Pierre},
  journal={Human Movement Science},
  volume={30},
  number={4},
  pages={783--791},
  year={2011},
  publisher={Elsevier},
  note={\url{https://doi.org/10.1016/j.humov.2010.08.008}}
}

@inproceedings{simonyan2014very,
  title={Very deep convolutional networks for large-scale image recognition},
  author={Simonyan, Karen and Zisserman, Andrew},
  year={2014},
  note={\url{arXiv:1409.1556}},
}

@inproceedings{deng2009imagenet,
  title={Imagenet: A large-scale hierarchical image database},
  author={Deng, Jia and Dong, Wei and Socher, Richard and Li, Li-Jia and Li, Kai and Fei-Fei, Li},
  booktitle={2009 IEEE Conference on Computer Vision and Pattern Recognition},
  pages={248--255},
  year={2009},
  organization={IEEE},
  note={\url{https://doi.org/10.1109/CVPR.2009.5206848}}
}

@inproceedings{lipton2015critical,
  title={A critical review of recurrent neural networks for sequence learning},
  author={Lipton, Zachary C and Berkowitz, John and Elkan, Charles},
  year={2015},
  note={\url{arXiv:1506.00019}},
}

@article{zhang2017drawing,
  title={Drawing and recognizing chinese characters with recurrent neural network},
  author={Zhang, Xu-Yao and Yin, Fei and Zhang, Yan-Ming and Liu, Cheng-Lin and Bengio, Yoshua},
  journal={IEEE Transactions on Pattern Analysis and Machine Intelligence},
  volume={40},
  number={4},
  pages={849--862},
  year={2017},
  publisher={IEEE},
  note={\url{https://doi.org/10.1109/TPAMI.2017.2695539}}
}

@inproceedings{you2016image,
  title={Image captioning with semantic attention},
  author={You, Quanzeng and Jin, Hailin and Wang, Zhaowen and Fang, Chen and Luo, Jiebo},
  booktitle={Proceedings of the IEEE Conference on Computer Vision and Pattern Recognition},
  pages={4651--4659},
  year={2016},
  note={\url{https://doi.org/10.1109/CVPR.2016.503}}
}

@article{impedovo2019handwriting,
  title={A handwriting-based protocol for assessing neurodegenerative dementia},
  author={Impedovo, Donato and Pirlo, Giuseppe and Vessio, Gennaro and Angelillo, Maria Teresa},
  journal={Cognitive Computation},
  volume={11},
  number={4},
  pages={576--586},
  year={2019},
  publisher={Springer},
  note={\url{https://doi.org/10.1007/s12559-019-09642-2}}
}

@article{teulings1991control,
  title={Control of stroke size, peak acceleration, and stroke duration in {P}arkinsonian handwriting},
  author={Teulings, Hans-Leo and Stelmach, George E},
  journal={Human Movement Science},
  volume={10},
  number={2-3},
  pages={315--334},
  year={1991},
  publisher={Elsevier},
  note={\url{https://doi.org/10.1016/0167-9457(91)90010-U}}
}

@article{eichhorn1996computational,
  title={Computational analysis of open loop handwriting movements in {P}arkinson's disease: a rapid method to detect dopamimetic effects},
  author={Eichhorn, TE and Gasser, T and Mai, N and Marquardt, C and Arnold, G and Schwarz, J and Oertel, WH},
  journal={Movement Disorders: Official Journal of the Movement Disorder Society},
  volume={11},
  number={3},
  pages={289--297},
  year={1996},
  publisher={Wiley Online Library},
  note={\url{https://doi.org/10.1002/mds.870110313}}
}

@article{zham2017distinguishing,
  title={Distinguishing different stages of {P}arkinson's disease using composite index of speed and pen-pressure of sketching a spiral},
  author={Zham, Poonam and Kumar, Dinesh K and Dabnichki, Peter and Poosapadi Arjunan, Sridhar and Raghav, Sanjay},
  journal={Frontiers in Neurology},
  volume={8},
  pages={435},
  year={2017},
  publisher={Frontiers},
  note={\url{https://doi.org/10.3389/fneur.2017.00435}}
}

@book{hastie2009elements,
  title={The elements of statistical learning: data mining, inference, and prediction},
  author={Hastie, Trevor and Tibshirani, Robert and Friedman, Jerome},
  year={2009},
  publisher={Springer Science \& Business Media},
  note={\url{https://doi.org/10.1007/978-0-387-84858-7}}
}







\end{document}